\PassOptionsToPackage{table}{xcolor}

\documentclass{article} 
\usepackage{iclr2027_conference,times}

\usepackage{amsmath,amsfonts,bm}

\def\eqref#1{equation~\ref{#1}}

\def\1{\bm{1}}

\DeclareMathAlphabet{\mathsfit}{\encodingdefault}{\sfdefault}{m}{sl}
\SetMathAlphabet{\mathsfit}{bold}{\encodingdefault}{\sfdefault}{bx}{n}

\newcommand{\Mbase}{M_{\mathrm{base}}}
\newcommand{\Mreal}{M_{\mathrm{real}}}
\newcommand{\Msyn}{M_{\mathrm{syn}}}

\usepackage{hyperref}
\usepackage{url}
\usepackage{graphicx}
\usepackage{amssymb}
\usepackage{booktabs}
\usepackage{capt-of}
\usepackage[font=small]{caption}
\usepackage{wrapfig}
\usepackage{float}
\usepackage{etoc}
\usepackage{xcolor}
\newcommand{\gr}[1]{\textcolor{gray}{#1}}

\title{ReGain: Restoring Subject Fidelity in Personalization on Synthetic Images} 

\author{Shubhang Bhatnagar \quad Ishan Bhatnagar \quad Viraj Shah \quad Narendra Ahuja \\
University of Illinois Urbana-Champaign  \\
\texttt{\{sb56, vjshah3, n-ahuja\}@illinois.edu, ishanb98@gmail.com}
}

\iclrfinalcopy 
\begin{document}

\maketitle
\lhead{Preprint}
\suppressfloats[t]
\etocdepthtag.toc{mtmain}

\begin{abstract}

Text-to-image diffusion models are personalized to a subject by DreamBooth fine-tuning on a handful of its images. Increasingly, these images come from a diffusion model rather than a camera. We show that fine-tuning on such synthetic images degrades subject fidelity, producing oversaturated color and excess high-frequency detail. To isolate the cause, we fine-tune two models from the same base model with the same DreamBooth recipe, one on real photos of a subject and one on synthetic images of that subject generated by the first. We trace the degradation to classifier-free guidance (CFG). For the model personalized on synthetic images, the angle between the conditional and unconditional noise predictions, and with it the norm of their difference, is much larger than for the model personalized on real photos. This inflation grows toward high frequencies and also appears at other prompts semantically close to the subject, such as its class noun, but not at unrelated ones. We propose ReGain, a training-free correction applied at sampling time that measures how much each frequency band of the guidance is inflated relative to the base model and scales that band down accordingly. ReGain needs no real photos. On Stable Diffusion v1.5, ReGain closes 51–64\% of the subject-fidelity gap to the model personalized on real photos, as measured by DINO, DINOv2 and CLIP-I. It also improves subject fidelity on SDXL and SD 3.5 and preserves text alignment on all three backbones.

\end{abstract}

\section{Introduction}

\begin{figure}[t]
\centering
\includegraphics[width=\linewidth]{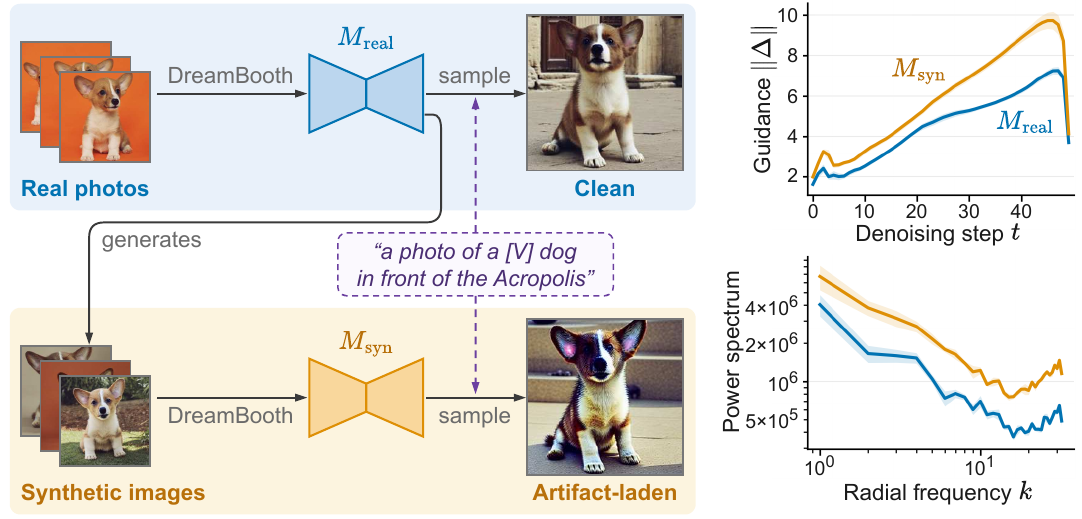}
\caption{\textbf{Personalizing a model on synthetic images of a subject degrades its fidelity.} Left: $\Mreal$ is DreamBooth fine-tuned from the base model $\Mbase$ on real photographs of the subject, and $\Msyn$ is fine-tuned from the same base with an identical recipe on images generated by $\Mreal$. Both are sampled with the same prompt and the same 10 seeds.
Right: compared to $\Mreal$, $\Msyn$ shows an inflated per-step $\|\Delta\|$ across denoising steps and an inflated latent power spectrum in its generated images.}
\label{fig:teaser}
\vspace{-8pt}
\end{figure}
Text-to-image diffusion models \citep{rombach2022high} are personalized to a user's subject so that text prompts can place it in novel scenes on demand. DreamBooth style  \citep{ruiz2023dreambooth} personalization methods fine-tune a diffusion model on a handful of subject images to associate it with a unique identifier.  Increasingly, however, the subject images themselves come from a generative model rather than a camera, e.g. as edited variants of a real subject. 

We study this change of data provenance under controlled conditions: $\Mreal$ is personalized with DreamBooth \citep{ruiz2023dreambooth} on real photographs, and $\Msyn$ is fine-tuned from the same base with an identical recipe on images generated by $\Mreal$, so any difference in their outputs is due to the training data alone. With the same prompt and seeds, $\Mreal$ produces a clean image, while $\Msyn$'s is oversaturated and overloaded with high-frequency texture as seen in Fig.~\ref{fig:teaser} (left). Its output latents carry more power at every frequency, and the excess grows in the high bands (Fig.~\ref{fig:teaser}, top right). We trace this back to guidance. Classifier-free guidance (CFG) \citep{ho2022classifier} steers each denoising step along $\Delta = \epsilon_c - \epsilon_{\varnothing}$, and for $\Msyn$, $\|\Delta\|$ is inflated throughout denoising, with a gap that widens as sampling proceeds (Fig.~\ref{fig:teaser}, bottom right).


\textbf{Findings.} The $\Delta$ inflation has four properties.
\emph{(1) Angle-driven.} The norms of $\epsilon_c$ and $\epsilon_{\varnothing}$ do not grow, but the angle between them exceeds $\Mreal$'s at every step so their vector difference $\|\Delta\|$ inflates
\emph{(2) Semantic.} Prompts without the identifier, using only the subject's class name (e.g., ``a dog''), show similar inflation, and so do related classes (e.g. cats) never seen in fine-tuning. Unrelated prompts show none.
\emph{(3) A property of the model.} The inflation also appears away from $\Msyn$'s sampling trajectories, when it denoises noisy copies of its own training images.
\emph{(4) High-frequency and time-varying.} High-frequency bands inflate several-fold over the early-to-middle trajectory, far more than low-frequency bands. Each band follows its own time profile. A lower guidance scale acts uniformly across bands and steps, so it cannot undo this.

\textbf{Method.} Based on these findings we propose ReGain, a training-free
correction applied at sampling time, so it also repairs already-trained
checkpoints. Correcting the inflation requires a reference for how large $\Delta$ should be. $\Mreal$ could help, but it is available only in our analysis experiments, not in practical settings. The base model that
$\Msyn$ was fine-tuned from, however, is freely available, and its $\Delta$ for
the subject's class is uninflated. Because the inflation appears away from sampling (finding 3), ReGain measures it once, before sampling, it runs $\Msyn$ and the base model on noisy copies of $\Msyn$'s training images and records how much larger $\Msyn$'s $\Delta$ is in each frequency band and denoising step.
At sampling time, it lowers the guidance of each band and step by that factor (findings 1 and 4), where the subject tokens govern the prediction (finding 2), with plain CFG elsewhere.


\textbf{Contributions.}
\begin{itemize}
     \item We show that DreamBooth fine-tuning on synthetic subject images inflates the CFG guidance term $\Delta$, and characterize this inflation in four findings.
    \item Using these findings, we propose ReGain, a training-free correction that lowers the guidance per frequency band and denoising step, using the public base model as the reference. It needs no real photos, retraining or per-subject tuning.
    \item We evaluate ReGain on the $30$ subjects of the DreamBooth dataset with DreamBooth and DreamBooth-LoRA on SD 1.5 and with DreamBooth-LoRA on SDXL and SD~3.5. In all four settings, ReGain significantly improves subject fidelity (DINO, DINOv2 and CLIP-I) while preserving prompt fidelity (CLIP-T). On SD 1.5, it closes $51$--$64\%$ of the subject-fidelity gap to a model personalized on real photos.

\end{itemize}

\section{Related Work}
\label{sec:related}

\textbf{Subject personalization.} Perosnalization methods may bind a subject to a
unique identifier by training on a handful of its images
\citep{ruiz2023dreambooth, gal2023image}, updating all weights, selected layers
\citep{kumari2023multi}, or low-rank adapters \citep{hu2022lora, shah2024ziplora},
optionally with drift regularization \citep{lee2024direct, kim2026preserve}. Tuning-free methods condition on the subject images
through encoders or adapters \citep{ye2023ip, tan2025ominicontrol, wu2025less}
but reproduce fine identity detail less faithfully, so fine-tuning remains the
choice when fidelity matters.

\textbf{Training on synthetic data.}   Subject images used for personalization
are increasingly produced by generative models \citep{avrahami2024chosen,
kumari2025generating, li2025iccustom}.   Generative models retrained on their own
outputs over several generations lose diversity and fidelity \citep{shumailov2024ai, alemohammad2024self, bertrand2024stability}, and proposed remedies mix real data back into training \citep{gerstgrasser2024accumulate} or guide sampling away from a model trained on self-generated data \citep{alemohammad2025self}.
\citet{yoon2025model} identify the CFG guidance scale as a driver of collapse over many generations of training.  In contrast to these works, our work studies (1) a single round of DreamBooth fine-tuning, not generic training (without a subject token) of a chain of models, and (2) in our case we show that the CFG guidance term (not the scale) itself is miscalibrated as the angle between conditional and unconditional predictions increases, inflating guidance in a band-specific way, and correct it without training.

\textbf{Guidance analyses and fixes.} Classifier-free guidance \citep{ho2022classifier} is the standard steering mechanism of text-to-image diffusion \citep{rombach2022high}, and many works adjust when, where, or how strongly it is applied \citep{kynkaanniemi2024applying, sadat2024cads, lin2024common, sadat2025eliminating, karras2024guiding, shen2024rethinking, chung2025cfgpp}, including per frequency band \citep{sadat2025fdg} and for personalized models \citep{chan2024subjectagnostic, park2025steering, jeong2025mindiff}. These works do not consider the inflation of the guidance term itself caused by model personalization on synthetic images, which ReGain corrects before any of them are applied.



\section{Method}
\label{sec:method}

\subsection{Setup and notation}
\label{sec:setting}
To isolate what synthetic training data does to the guidance, we compare two
models trained from the same base model $\Mbase$ by the DreamBooth
objective \citep{ruiz2023dreambooth} (Figure~\ref{fig:overview}A). The first,
$\Mreal$, is fine-tuned on real photographs of a subject, binding it
to a unique identifier [V]. As a running example we use subject dog6 of the
DreamBooth dataset on SD 1.5. Its subject prompt
$c_{\mathrm{subj}}$, \emph{``a photo of a [V] dog''}, names the subject through
[V], and its class prompt $c_{\mathrm{class}}$, \emph{``a photo of a dog''}, omits it. The second model, $\Msyn$, is fine-tuned by the same recipe on synthetic images of the subject generated by $\Mreal$. Because the two models share the base weights and the training recipe, any difference in their guidance traces to the training images alone. The two training sets are comparably diverse (see Appendix~\ref{app:diversity}), so what distinguishes them is the origin of their images . In practice, only $\Msyn$, its training images, and $\Mbase$ are available. $\Mreal$ is used solely for the analysis in Section~\ref{sec:findings} and for evaluation.

At step $t$ (counted from the noisiest state, $t = 0$), the noisy latent $z_t$ has signal and noise variances $\bar\alpha_t$ and $1 - \bar\alpha_t$. Classifier-free guidance \citep{ho2022classifier}  combines predictions under the prompt $c$ and the null prompt $\varnothing$:
\begin{equation}
\epsilon_w \;=\; \epsilon_{\varnothing} + w\,(\epsilon_c - \epsilon_{\varnothing}),
\qquad
\Delta \;=\; \epsilon_c - \epsilon_{\varnothing} ,
\label{eq:cfg}
\end{equation}
with guidance weight $w$. Superscripts denote the source model, e.g.\ $\Delta^{\Mreal}$ is the guidance of $\Mreal$.

\subsection{Why fine-tuning on synthetic images inflates guidance}

We compare the noise predictions of $\Msyn$ and $\Mreal$ at every step as both
sample $c_{\mathrm{subj}}$ from the same $10$ initial noises with the CFG of
\eqref{eq:cfg}. The guidance term $\Delta$ of $\Msyn$ is systematically larger
in norm than that of $\Mreal$, an excess we call the $\Delta$~inflation. Its norm is controlled by three quantities: the unconditional norm
$r = \|\epsilon_{\varnothing}\|$, the ratio
$\delta = \|\epsilon_c\| / \|\epsilon_{\varnothing}\|$, and the angle $\theta$
between $\epsilon_{\varnothing}$ and $\epsilon_c$:
\begin{equation}
\|\Delta\|^2 \;=\; r^2\big(1 + \delta^2 - 2\delta\cos\theta\big), \qquad
\theta = \arccos\!\big( \langle \epsilon_{\varnothing}, \epsilon_c \rangle \,/\,
                        \|\epsilon_{\varnothing}\| \|\epsilon_c\| \big) .
\label{eq:scalars}
\end{equation}
We report these measurements on one DreamBooth subject, dog6, and repeat them on all $30$ subjects in Appendix~\ref{app:all-subjects}.

\begin{figure}[t]
\centering
\includegraphics[width=\linewidth]{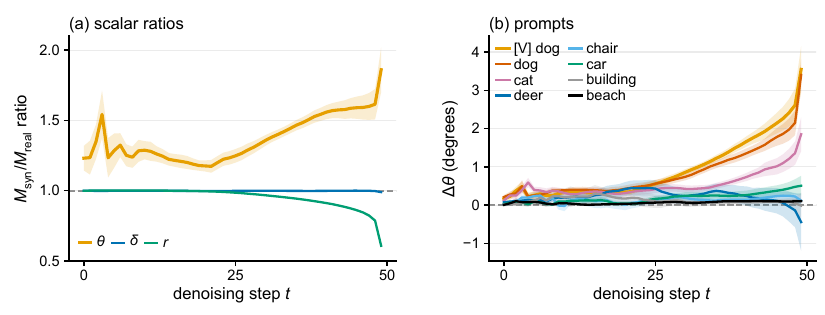}
\caption{(a)$\Msyn/\Mreal$ ratio of $\theta$, $\delta$ and $r$. (b)~Angle excess $\theta^{\Msyn}(t) - \theta^{\Mreal}(t)$ for the subject prompt and increasingly semantically distant prompts. 10
seeds mean.}
\label{fig:findings}
\end{figure}

\textbf{(1) The $\Delta$ inflation is angle-driven.} Of the three scalars, only the angle raises $\|\Delta\|$ as seen in Figure~\ref{fig:findings} (a). $\theta^{\Msyn}$ exceeds $\theta^{\Mreal}$ at every step, by $37\%$ on average over the $50$ steps. $\delta$ stays within $1\%$ of unity for both models, and $r^{\Msyn}$ stays within $3\%$ of $r^{\Mreal}$ until step $30$ and then falls below it, which works against the inflation. Rewriting \eqref{eq:scalars} as $\|\Delta\|^2 = r^2\big[(1-\delta)^2 + 4\delta\sin^2(\theta/2)\big]$, $\delta \approx 1$ gives $\|\Delta\| \approx 2r\sin(\theta/2) \approx r\theta$, so the guidance norm follows the angle: $\|\Delta^{\Msyn}\|$ is larger at every step, by $29\%$ on average. Repeated on all $30$ DreamBooth subjects in Appendix~\ref{app:all-subjects}, the excess averages $48\%$. Since CFG adds $w\Delta$ to $\epsilon_{\varnothing}$, $\Msyn$ is over-guided at the same $w$.

\textbf{(2) The $\Delta$ inflation follows the subject's semantic neighborhood.} In prompt space,
the angle excess $\theta^{\Msyn}(t) - \theta^{\Mreal}(t)$ (mean over seeds) for
prompts at increasing semantic distance from the identifier splits into two
tiers (Figure~\ref{fig:findings}b). The class noun (\emph{``a photo of
a dog''}) inherits nearly the full inflation, and \emph{cat}, an animal never seen
in fine-tuning, about half of it. Prompts farther away (\emph{deer}, \emph{chair}, \emph{car}, \emph{building},
\emph{beach}) show little to no inflation until the final steps.

\textbf{(3) The $\Delta$ inflation persists when both models see the same input.}  Findings 1 and 2 compare the two models on their own sampling trajectories, which differ. Toward a fix, we test whether the inflation persists when both models are evaluated at the same inputs, namely forward-noised training images of $\Msyn$, $z_t = \sqrt{\bar\alpha_t}\, z_0 + \sqrt{1 - \bar\alpha_t}\, n$, with $z_0$ the latent encoding of a training image and $n \sim \mathcal{N}(0, I)$ drawn afresh at every step. There, where $\Msyn$ should be most faithful, the guidance term of $\Msyn$ still exceeds that of $\Mreal$ in band energy. Retraining $\Msyn$ on images sampled from $\Mreal$ at lower guidance weights leaves the inflation in place (dog6, Appendix~\ref{app:gen-guidance}). Adding the prior-preservation loss of DreamBooth to fine-tuning does not remove it either (Appendix~\ref{app:prior-pres}).
\label{sec:findings}

\begin{figure}[b]
\centering
\includegraphics[width=\linewidth]{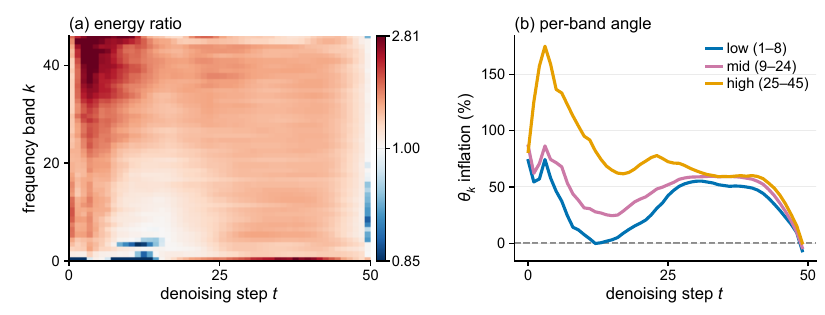}
\caption{(a)~$\Msyn/\Mreal$ band-energy ratio of $\Delta$ per frequency band $k$ and step $t$; red marks inflation. (b)~$\theta$ inflation of the low, mid and high bands over steps. 10
seeds mean.}
\label{fig:findings-bands}
\end{figure}

\textbf{(4) The $\Delta$ inflation concentrates in high frequencies and varies
over time.} We next ask whether the inflation is uniform across frequencies and steps, since neural networks learn different frequencies at different rates \citep{rahaman2019spectral}. We split $\Delta$ into $K$ frequency bands, from the constant component ($k = 0$) to the finest detail ($k = K - 1$), with $B_k(\Delta)$ denoting the component of $\Delta$ in band $k$. For analysis we group the bands into low, mid and high (boundaries in Appendix~\ref{app:impl}). Both models are evaluated at the states $\Msyn$ visits during sampling. To measure the inflation in each band, we compare the band energies of the two guidance terms, averaged over seeds, as $\sqrt{\|B_k(\Delta^{\Msyn})\|^2 / \|B_k(\Delta^{\Mreal})\|^2}$.  Figure~\ref{fig:findings-bands}a shows this ratio for every band and step. The inflation grows from the low to the high bands and peaks earliest in the high bands as seen in Figure~\ref{fig:findings-bands} (a, b). Each band has its own time profile, so no single guidance weight describes the inflation.


\subsection{ReGain: restoring the guidance from the base model}
\label{sec:regain}

\begin{figure}[t]
\centering
\includegraphics[width=\linewidth]{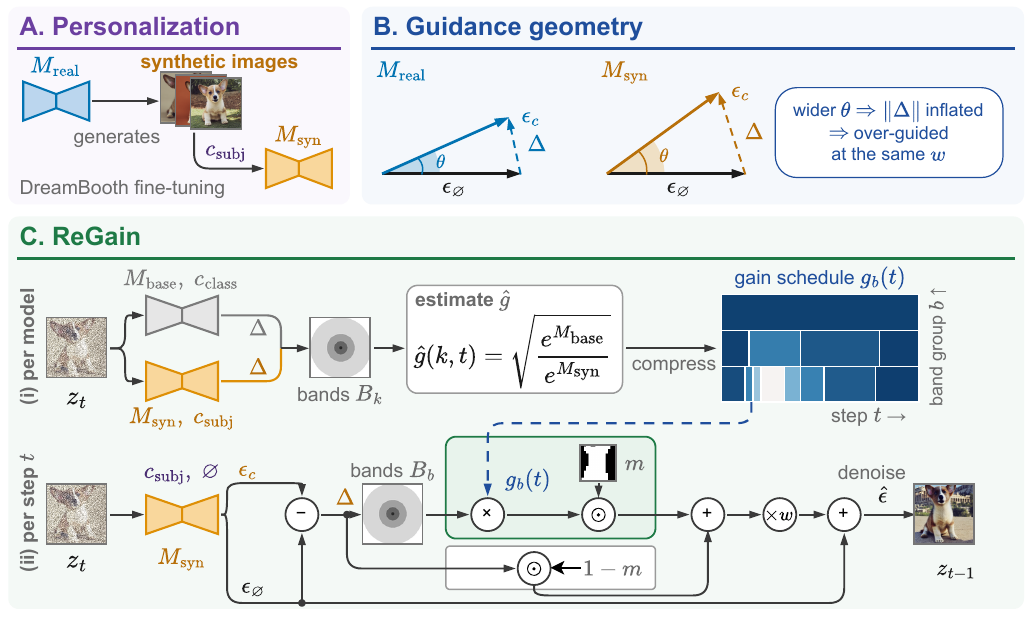}
\caption{\textbf{Overview.}
(A)~$\Msyn$ is DreamBooth fine-tuned on synthetic subject images generated by
$\Mreal$.
(B)~The angle $\theta$ between the unconditional and conditional predictions
is wider for $\Msyn$, so its guidance term $\Delta = \epsilon_c -
\epsilon_{\varnothing}$ is inflated.
(C)~ReGain. (i)~Once per model, $\Mbase$ and $\Msyn$ are evaluated at
forward-noised training images $z_t$; the square root of the ratio of their
masked band energies $e$ gives the gain $\hat g(k,t)$, compressed into the
schedule $g_b(t)$ (darker blue attenuates more; Figure~\ref{fig:regain}).
(ii)~At each step, $\Delta$ is rescaled per band group by $g_b(t)$ inside the
subject mask $m$ (green) and left unchanged outside it (grey), then scaled by
$w$ and added to $\epsilon_{\varnothing}$, \eqref{eq:masked-update}.}
\label{fig:overview}
\end{figure}

To correct the inflation at sampling time, ReGain needs a reference for how large $\Delta$ should be at each band and step. We take this reference from the base model $\Mbase$, which never saw the synthetic images. Concretely, ReGain has two parts as shown in Figure~\ref{fig:overview}C : (1) a one-time calibration, which estimates how much $\Msyn$'s guidance exceeds $\Mbase$'s in each band and step, as a gain $g(k,t)$, and quantizes it into a compact schedule; and (2) a correction at sampling time, which rescales each band of $\Delta$ by the scheduled gain inside the subject mask. Neither part involves training.

\textbf{Measuring the $\Delta$ inflation against the base model.} To measure how much $\Msyn$'s guidance exceeds that of the base model, we evaluate both at the forward-noised training images of finding~3 with the same noise draws: $\Mbase$ under $c_{\mathrm{class}}$, since it has never seen the identifier, and $\Msyn$ under $c_{\mathrm{subj}}$. At each band $k$ and step $t$, the masked band energy of
the guidance term is given by
\begin{equation}
e(k,t) \;=\; \frac{ \big\| \, m \odot B_k( \Delta ) \, \big\|^2 }{ |m| \, C } ,
\label{eq:estat}
\end{equation}

where $m$ is a binary subject mask and $|m|$ and $C$ are the numbers of masked latent pixels and latent channels. The mask is read from the attention maps of $\Msyn$ at the same state, by the segmentation stage of S-CFG \citep{shen2024rethinking} on U-Net models and by Seg4Diff \citep{kim2025seg4diff} on transformer models, and is shared by both models so that their energies are compared over the same pixels. Appendix~\ref{app:mask} details its construction.  We define the gain as the square root of the ratio of their means,
\begin{equation}
\hat{g}(k,t) \;=\;
\sqrt{ \;
\mathbb{E}\big[ e^{\Mbase}(k,t) \big] \;\big/\;
\mathbb{E}\big[ e^{\Msyn}(k,t) \big] \; } .
\label{eq:estimator}
\end{equation}

\begin{wrapfigure}{r}{0.46\textwidth}
\vspace{-\baselineskip}
\centering
\includegraphics[width=\linewidth]{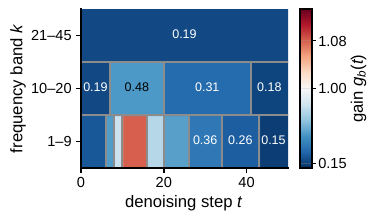}
\caption{Gain schedule $g_b(t)$ for dog6 from the base model (\eqref{eq:estimator}),
one row per band group $b$.}
\label{fig:regain}
\vspace{-\baselineskip}
\end{wrapfigure}
\textbf{The gain schedule.} The estimated gain $\hat g(k,t)$ has one entry per band and step, each computed from only a few training images and noise draws, so it is noisy at this resolution. To average out this noise and obtain a compact schedule, we compress it into a small number of rectangular cells of constant gain over contiguous bands and steps. The compression proceeds in two stages:
the bands are first cut into the fewest contiguous band groups $b$, and the steps of each group are then cut into the fewest contiguous segments, such that at each stage the cells deviate from $\hat g$ by at most a tolerance $\tau$ in root
mean square. Each stage is solved exactly by a one-dimensional dynamic program
on the squared error (tolerance in Section~\ref{sec:setup}). The DC band is exempt from this compression and keeps its measured gain. Figure~\ref{fig:regain} shows the resulting schedule for the running dog6 example.

\textbf{Correcting at sampling time.}
Because the inflation varies across bands (finding~4) and is tied to the subject (finding~2), the correction rescales each band
group separately and acts only where the subject tokens govern the prediction,
a choice the ablation of Table~\ref{tab:ablation} validates. The bands
partition the spectrum, so $\Delta = \sum_b B_b(\Delta)$ exactly for any
grouping of the $K$ bands into contiguous band groups $b$, with
$B_b = \sum_{k \in b} B_k$. Rescaling each group by a gain $g_b(t)$ and
applying the result inside a binary subject mask $m$, with plain CFG outside,
gives the corrected prediction at guidance weight $w$:
\begin{equation}
\hat{\epsilon} \;=\; \epsilon_{\varnothing}
\;+\; w \, m \odot \sum_b g_b(t)\, B_b(\Delta)
\;+\; w \, (1 - m) \odot \Delta ,
\label{eq:masked-update}
\end{equation}
where $\odot$ is the elementwise product. The band projections $B_b(\Delta)$
are computed on the full guidance term, and the mask is applied to the result,
so that each band is rescaled globally before the spatial split. At sampling, $m$ is read at each step from $\Msyn$ under the sampled prompt, whose scene words serve as the background tokens; Appendix~\ref{app:mask} details both constructions. When every $g_b = 1$, \eqref{eq:masked-update} reduces to plain CFG.



\section{Experiments}
\label{sec:experiments}

\subsection{Setup}
\label{sec:setup}

\textbf{Dataset.} We evaluate on the DreamBooth dataset
\citep{ruiz2023dreambooth}, $30$ subjects with $4$--$6$ real photographs
each, $25$ evaluation prompts per subject, $3$ seeds, and one
image per prompt and seed.

\textbf{Baselines.} For every subject we build the $\Mreal$--$\Msyn$ pair of
Section~\ref{sec:setting}, with $\Msyn$ fine-tuned on five images generated by
$\Mreal$, in four settings: DreamBooth and DreamBooth-LoRA \citep{hu2022lora} on
Stable Diffusion v1.5 (SD1.5) \citep{rombach2022high}, and DreamBooth-LoRA on SDXL
\citep{podell2024sdxl} and Stable Diffusion 3.5 (SD3.5) \citep{esser2024scaling},
each over all $30$ subjects. In each setting we compare $\Msyn$ with plain CFG
against $\Msyn$ with ReGain, sampled at the same seed, prompt, sampler and
guidance weight $w$, with the gain schedule estimated once per subject from
its five training images, and report $\Mreal$ as the reference. On SD1.5 we
also compare with three guidance methods applied to $\Msyn$. S-CFG
\citep{shen2024rethinking} rescales the guidance per semantic region, CFG++
\citep{chung2025cfgpp} renoises each step with the unconditional prediction,
and FDG \citep{sadat2025fdg} guides the low and high frequencies with separate
weights. Each runs with its authors' default settings at the same seed, prompt
and number of sampling steps. All training and sampling hyperparameters,
including those of the baselines, are listed in Appendix~\ref{app:impl}.

\textbf{Subject mask.} The mask $m$ of Section~\ref{sec:regain} is rebuilt at
every sampling step from $\Msyn$'s attention under the prompt being sampled.
Its construction at calibration, sampling, and visualization are in
Appendix~\ref{app:mask}.

\textbf{Metrics.} Subject fidelity is the cosine similarity between
embeddings of a generated image and the subject's real photographs, with
CLIP ViT-B/32 (CLIP-I), DINO ViT-S/16 (DINO) and DINOv2 ViT-S/14 (DINOv2)
\citep{radford2021learning, caron2021emerging, oquab2024dinov2}. Prompt
fidelity (CLIP-T) is the cosine similarity between the CLIP embeddings of
the image and of the prompt with the identifier removed. Over-guidance
artifacts are measured by mean saturation, root-mean-square contrast and the
high-band energy share defined in Section~\ref{sec:artifacts}.

\textbf{Implementation details.} All models except SD3.5 are sampled with DDIM
\citep{song2021ddim} for $T = 50$ steps at guidance weight $w = 7.5$, and SD3.5
with its flow-matching Euler sampler for $40$ steps at $w = 7.0$. The gain
schedule is compressed with tolerance $\tau = 0.05$. The measurement setup of Section~\ref{sec:findings}, the frequency bands and the full fine-tuning and sampling settings are in Appendix~\ref{app:impl}.

\subsection{Main comparison}
\label{sec:results-single}

\begin{figure}[t]
\centering
\includegraphics[width=\linewidth]{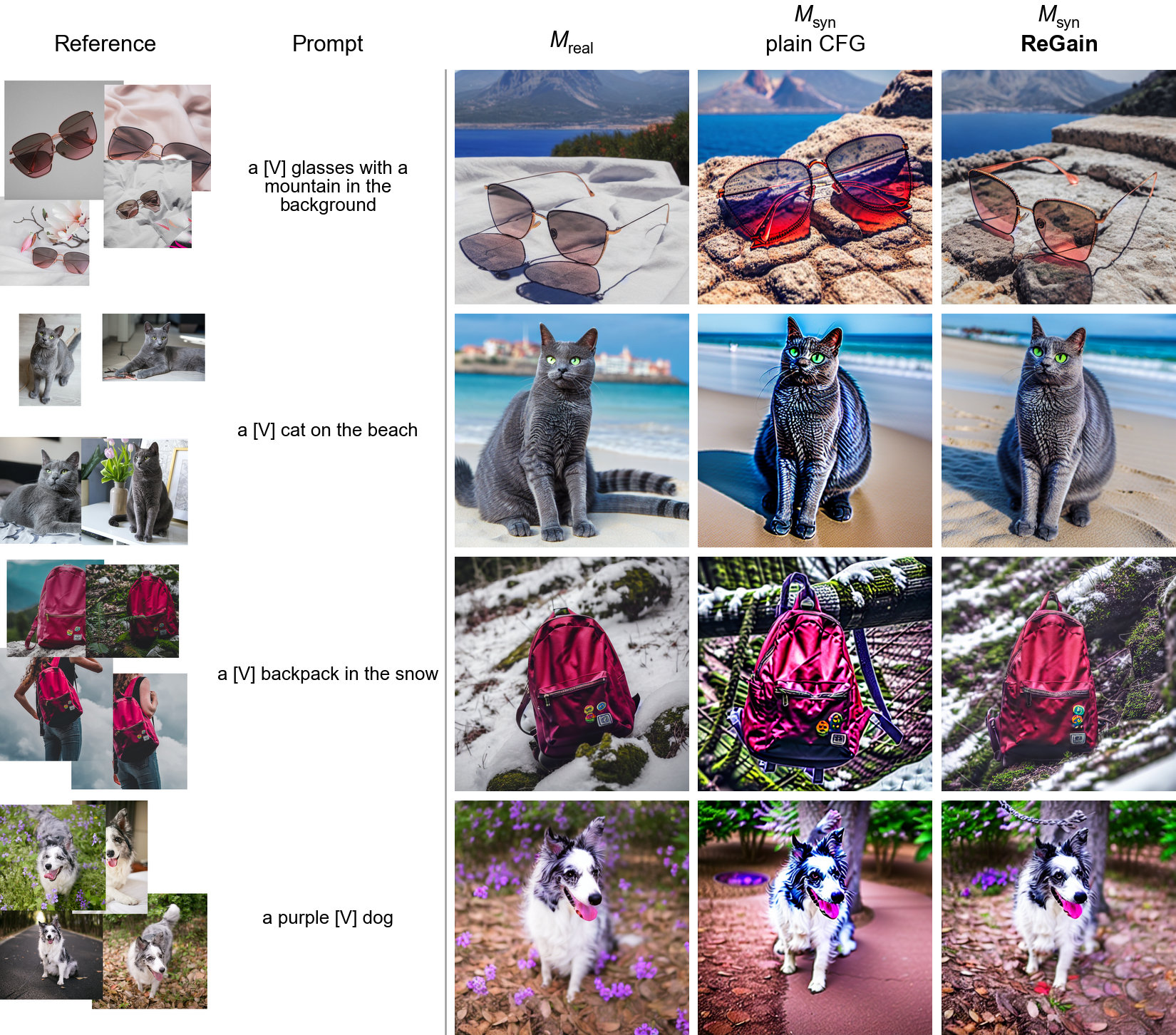}
\caption{Qualitative comparison on SD1.5 with DreamBooth. Each
row is one subject, prompt and seed shared by the three models. More
subjects and settings are in Appendix~\ref{app:more-results}.}
\label{fig:results-single}
\vspace{-10pt}
\end{figure}

\begin{table}[t]
\centering
\footnotesize
\setlength{\tabcolsep}{6pt}
\captionof{table}{Subject and prompt fidelity on DreamBooth, averaged over
$30$ subjects, $25$ prompts and $3$ seeds (higher is better). The first row is
the reference model $\Mreal$, and every other row samples from $\Msyn$.
Baselines, on SD1.5 only: S-CFG~\citep{shen2024rethinking},
CFG++~\citep{chung2025cfgpp} and FDG~\citep{sadat2025fdg}. Bold marks the best
$\Msyn$ result. Gap recovered is the part of the drop from $\Mreal$ to $\Msyn$,
both with CFG, that ReGain wins back. For CLIP-T we report the change from
$\Msyn$ + CFG instead.}
\label{tab:metrics}
\begin{tabular*}{\linewidth}
{lcccccccc@{}}
\toprule
& \multicolumn{4}{c}{\makebox[0pt]{\emph{SD1.5, DreamBooth}}} & \multicolumn{4}{c}{\makebox[0pt]{\emph{SD1.5, DreamBooth-LoRA}}} \\
\cmidrule(lr){2-5}\cmidrule(lr){6-9}
Method & CLIP-I & DINO & DINOv2 & CLIP-T & CLIP-I & DINO & DINOv2 & CLIP-T \\
\midrule
\gr{$\Mreal$ + CFG} & \gr{0.814} & \gr{0.681} & \gr{0.661} & \gr{0.300} & \gr{0.791} & \gr{0.630} & \gr{0.606} & \gr{0.306} \\
$\Msyn$ + CFG       & 0.788 & 0.611 & 0.611 & 0.289 & 0.750 & 0.540 & 0.510 & 0.300 \\
\midrule
$\Msyn$ + S-CFG      & 0.781 & 0.605 & 0.604 & \textbf{0.292} & 0.744 & 0.536 & 0.503 & 0.302 \\
$\Msyn$ + CFG++      & 0.788 & 0.606 & 0.608 & 0.289 & 0.748 & 0.535 & 0.508 & 0.299 \\
$\Msyn$ + FDG        & 0.792 & 0.608 & 0.614 & 0.284 & 0.755 & 0.538 & 0.517 & 0.297 \\
\midrule
$\Msyn$ + CFG + ReGain & \textbf{0.802} & \textbf{0.647} & \textbf{0.643} & 0.290 & \textbf{0.775} & \textbf{0.594} & \textbf{0.568} & \textbf{0.304} \\
\rowcolor{teal!15}
\quad$\hookrightarrow$ Gap recovered & 54\% & 51\% & 64\% & $+0.001$ & 61\% & 60\% & 60\% & $+0.004$ \\
\midrule
& \multicolumn{4}{c}{\makebox[0pt]{\emph{SDXL, DreamBooth-LoRA}}} & \multicolumn{4}{c}{\makebox[0pt]{\emph{SD3.5, DreamBooth-LoRA}}} \\
\cmidrule(lr){2-5}\cmidrule(lr){6-9}
\gr{$\Mreal$ + CFG} & \gr{0.757} & \gr{0.558} & \gr{0.535} & \gr{0.289} & \gr{0.811} & \gr{0.682} & \gr{0.650} & \gr{0.311} \\
$\Msyn$ + CFG       & 0.724 & 0.477 & 0.467 & 0.278 & 0.776 & 0.586 & 0.567 & 0.307 \\
\midrule
$\Msyn$ + CFG + ReGain & \textbf{0.736} & \textbf{0.502} & \textbf{0.494} & \textbf{0.279} & \textbf{0.792} & \textbf{0.605} & \textbf{0.596} & \textbf{0.307} \\
\rowcolor{teal!15}
\quad$\hookrightarrow$ Gap recovered & 36\% & 31\% & 40\% & $+0.001$ & 46\% & 20\% & 35\% & $0.000$ \\
\bottomrule
\end{tabular*}
\vspace{-15pt}
\end{table}

\textbf{Quantitative results.} Table~\ref{tab:metrics} reports subject and
prompt fidelity in the four settings. Fine-tuning on model-generated images
costs subject fidelity on every metric and in every setting, and ReGain
recovers a consistent share of that loss at sampling time, with no
retraining: more than half of the gap to $\Mreal$ on SD1.5, and between a
fifth and a half on SDXL and SD3.5. Prompt fidelity is left intact: CLIP-T
changes by at most $0.004$ relative to plain CFG in every setting. None of the
three guidance baselines closes the gap. FDG recovers at most a sixth of it,
none of it on DINO, and lowers CLIP-T. S-CFG and CFG++ score at or below plain
CFG on all three subject fidelity metrics. The
over-guidance artifacts behind the fidelity loss, which these metrics capture
only indirectly, are measured in Section~\ref{sec:artifacts}.

\begin{figure}[h]
\centering
\includegraphics[width=\linewidth]{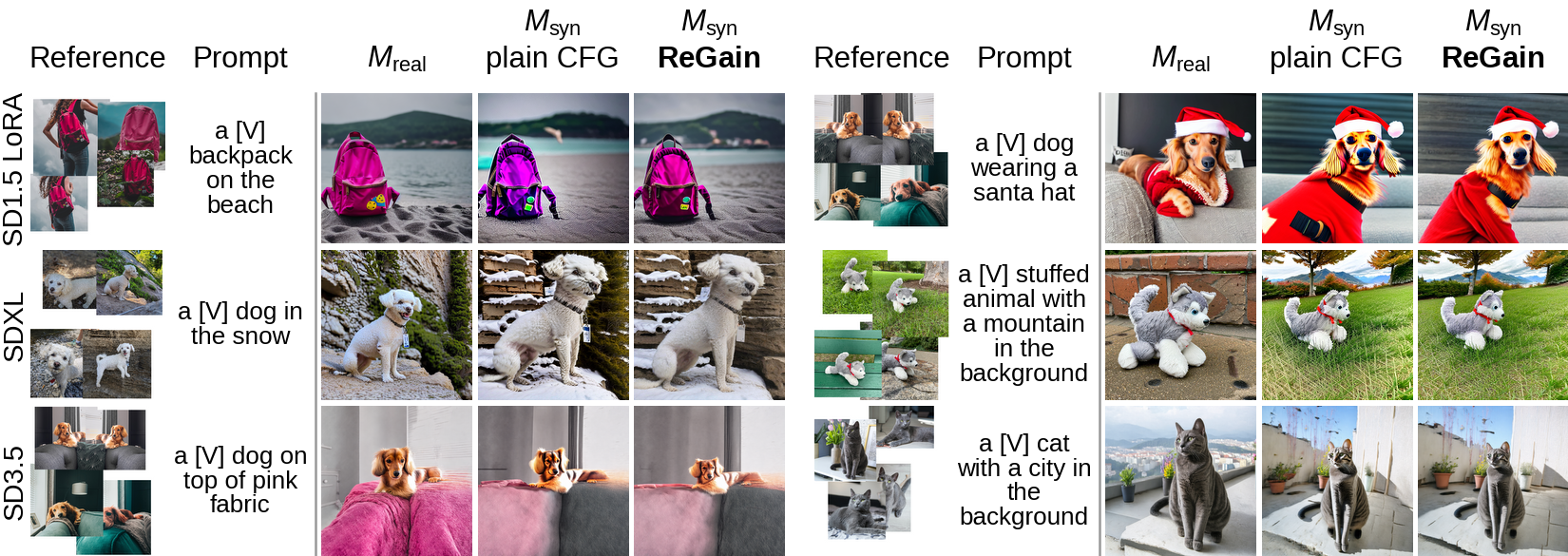}
\caption{Qualitative comparison across fine-tuning forms and base models. Each
row is one base model and each half one subject, prompt and seed shared by
the three models; nothing is retuned per base model. The SD3.5
row shows the chain of Table~\ref{tab:metrics}.}
\label{fig:results-backbone}
\vspace{-5pt}
\end{figure}

\textbf{Qualitative results.} Figure~\ref{fig:results-single} shows four
subjects, each at one prompt and seed shared by the three models. Under plain
CFG, $\Msyn$ renders the subject with oversaturated color, harsh contrast and
excess fine texture, the over-guidance artifacts of Figure~\ref{fig:teaser},
and the scene loses detail around it. ReGain, from the same model and seed,
brings the subject's color and texture back toward $\Mreal$'s while keeping the
prompt's scene, and the attribute change of the last column (the dog recolored
purple) is still carried out. Figure~\ref{fig:results-backbone} shows the same
behavior under DreamBooth-LoRA on SD1.5, on SDXL and on SD3.5.

\subsection{Over-guidance artifact metrics}
\label{sec:artifacts}

\begin{wraptable}{r}{0.51\textwidth}
\vspace{-\baselineskip}
\captionsetup{justification=raggedright,singlelinecheck=false}
\footnotesize
\setlength{\tabcolsep}{2pt}
\captionof{table}{Over-guidance artifacts inside the subject region (SD1.5,
DreamBooth). Bold marks the $\Msyn$ row closer to $\Mreal$.}
\label{tab:artifacts}
\begin{tabular}{@{}lccc@{}}
\toprule
Method & Saturation & Contrast & High band \\
\midrule
\gr{$\Mreal$ + CFG} & \gr{0.301} & \gr{0.211} & \gr{0.061} \\
$\Msyn$ + CFG            & 0.355 & 0.258 & 0.069 \\
$\Msyn$ + CFG + ReGain   & \textbf{0.309} & \textbf{0.234} & \textbf{0.066} \\
\rowcolor{teal!15}
$\hookrightarrow$ Gap recovered & 85\% & 51\% & 38\% \\
\bottomrule
\end{tabular}
\end{wraptable}
Over-guided samples carry a characteristic artifact signature: excess saturation
and contrast and inflated high-frequency content, the known symptoms of
over-guidance \citep{sadat2025eliminating} and the same signature reported for
models trained on their own high-CFG samples \citep{yoon2025model}. We quantify
it with three per-image statistics, mean HSV saturation, root-mean-square
grayscale contrast and the high-frequency fraction of the image's Fourier
power (definitions in Appendix~\ref{app:artifact-metrics}), and take $\Mreal$'s
value as the target, since the goal is to restore the reference model's image
statistics. This section evaluates the SD1.5 DreamBooth setting.

\textbf{Where they are measured.} The correction acts inside the subject mask
and leaves plain CFG outside it. We therefore report the three statistics
inside the mask that steered $\Msyn$, over identical pixels for all three
methods (Appendix~\ref{app:artifact-metrics}), and the full-frame values in the
text.

\textbf{Results.} Table~\ref{tab:artifacts} reports the three statistics
inside the subject region. Under plain CFG, $\Msyn$ is $18\%$ over-saturated,
$22\%$ over-contrasted and carries $13\%$ excess high-band power relative to
$\Mreal$: the over-guidance artifacts of Figure~\ref{fig:teaser}, and the
image-space counterpart of the spectral excess in Figure~\ref{fig:findings-bands}a.
ReGain moves all three statistics back toward $\Mreal$ and overshoots none of
them. Most of the saturation excess is removed, along with
about half of the contrast excess and a third of the high-band excess. On the full frame the same recoveries read $77\%$, $35\%$ and $5\%$, diluted by the background, which covers most of the frame and is generated under plain CFG in every method.

\subsection{Analysis and ablations}
\label{sec:analysis}

All experiments in this section use DreamBooth on SD1.5 with the $30$
subjects, prompts and seeds of Table~\ref{tab:metrics}.

\begin{table}[t]
\captionsetup{justification=raggedright,singlelinecheck=false}
\footnotesize
\setlength{\tabcolsep}{3pt}
\begin{minipage}[b]{0.515\linewidth}
\captionof{table}{Subject-agnostic alternatives and the subject-mask ablation
(SD v1.5, DreamBooth): a lower guidance weight $w$, APG
\citep{sadat2025eliminating}, and ReGain w/o mask, which applies the gain to
every pixel; $w{=}7.5$ unless stated. Saturation is inside the subject region; bold marks the $\Msyn$
row closest to $\Mreal$.}
\label{tab:ablation}
\begin{tabular*}{\linewidth}[b]{@{\extracolsep{\fill}}lccc@{}}
\toprule
Method & DINO & CLIP-T & Saturation \\
\midrule
\gr{$\Mreal$}      & \gr{0.681} & \gr{0.300} & \gr{0.301} \\
$\Msyn$, plain CFG             & 0.611 & 0.289 & 0.355 \\
$\Msyn$, plain CFG ($w{=}5.0$) & 0.632 & 0.286 & 0.326 \\
$\Msyn$, plain CFG ($w{=}3.0$) & 0.647 & 0.281 & 0.292 \\
$\Msyn$, APG                   & 0.641 & 0.283 & 0.240 \\
$\Msyn$, ReGain                & 0.647 & \textbf{0.290} & \textbf{0.309} \\
$\Msyn$, ReGain w/o mask       & \textbf{0.652} & 0.282 & 0.293 \\
\bottomrule
\end{tabular*}
\end{minipage}\hfill
\begin{minipage}[b]{0.455\linewidth}
\captionof{table}{$\Delta$ inflation at the calibration states before and
after the gain schedule, by band group and compression tolerance (RMS, target
$1$, mean $\pm$ std over $30$ subjects).}
\label{tab:inflation}
\begin{tabular*}{\linewidth}[b]{@{\extracolsep{\fill}}lccccc@{}}
\toprule
& & \multicolumn{4}{c}{After, by tolerance} \\
\cmidrule(lr){3-6}
Band & Before & 0.02 & \textbf{0.05} & 0.1 & 0.2 \\
\gr{cells} & & \gr{70} & \gr{18} & \gr{6} & \gr{2} \\
\midrule
Low $k$  & 4.70 $\pm$ 1.19 & 1.02 & \textbf{1.06} & 1.13 & 1.29 \\
Mid $k$  & 4.95 $\pm$ 1.19 & 1.01 & \textbf{1.05} & 1.16 & 1.35 \\
High $k$ & 5.76 $\pm$ 1.38 & 1.03 & \textbf{1.11} & 1.30 & 1.57 \\
All      & 5.31 $\pm$ 1.26 & 1.02 & \textbf{1.08} & 1.22 & 1.45 \\
\bottomrule
\end{tabular*}
\end{minipage}
\end{table}

\begin{figure}[t]
\centering
\includegraphics[width=\linewidth]{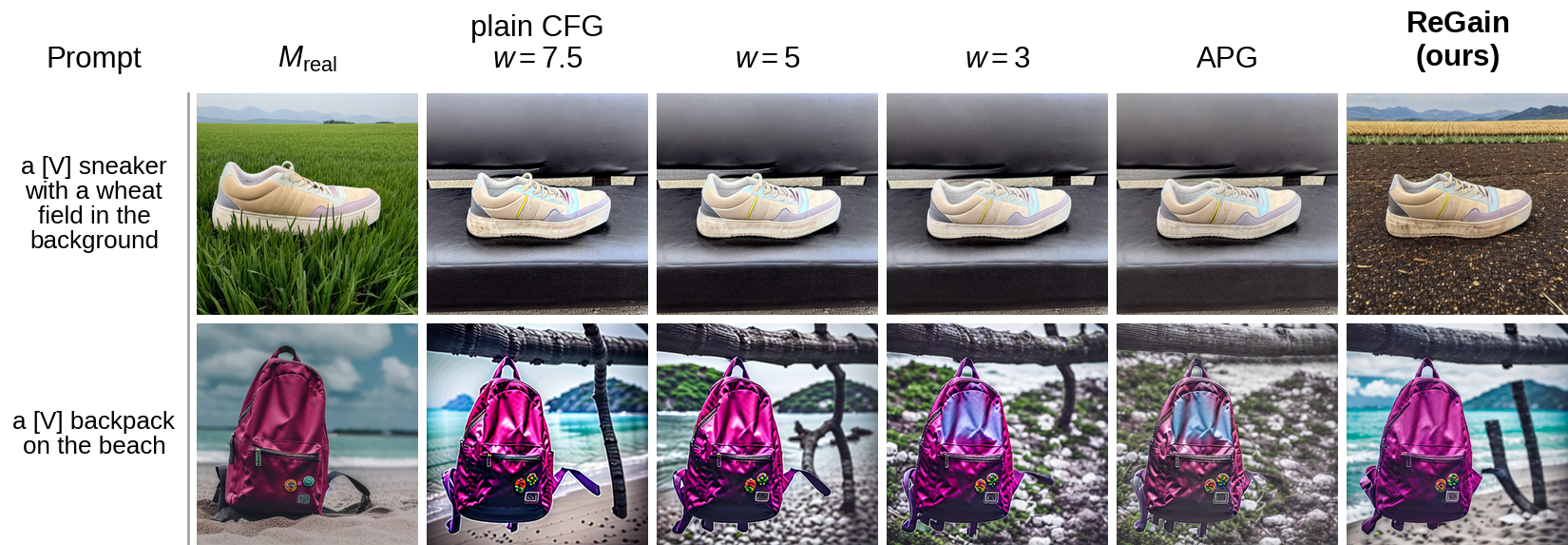}
\caption{Example generations from the baselines of
Table~\ref{tab:ablation}, all at the same seed. }
\label{fig:ablation-qual}
\vspace{-8pt}
\end{figure}

\textbf{Effect of subject-agnostic guidance changes.} The inflation is
structured in frequency and time and follows the subject tokens (findings 2
and 4, Section~\ref{sec:findings}), so no change applied alike to every band
and step should repair it. We test two such changes on $\Msyn$: lowering the
guidance weight to $w \in \{5.0, 3.0\}$, and adaptive projected guidance (APG)
\citep{sadat2025eliminating}, which removes the component of the guidance term
parallel to the conditional prediction. As seen in Table~\ref{tab:ablation}
and Figure~\ref{fig:ablation-qual}, (1) lowering $w$ removes guidance from the
scene as well as from the subject: CLIP-T falls by $0.003$ at $w{=}5.0$ and by
$0.008$ at $w{=}3.0$, the prompt's accessory or setting fades first, and
neither weight recovers DINO; (2) APG over-corrects: saturation drops to
$0.240$, well below $\Mreal$'s $0.301$, the frame is washed out, and CLIP-T
falls by $0.006$. ReGain is the only method that moves all three metrics
toward $\Mreal$ without overshooting any of them.

\textbf{Effect of the gain schedule.} Table~\ref{tab:inflation} measures the
excess guidance energy the schedule is built to remove: the square root of the
ratio of $\Msyn$'s masked band energy \eqref{eq:estat} to the base model's at
the same forward-noised training images, the inverse of the gain
\eqref{eq:estimator}, so that values above $1$ are inflation. As seen in
Table~\ref{tab:inflation}, (1) before correction the inflation rises from the
low-$k$ to the high-$k$ group, the ordering of the per-band angle inflation of
Section~\ref{sec:findings}, and (2) after the gain schedule at $\tau = 0.05$ it
lies within $10\%$ of $1$ on average (residual by step in
Appendix~\ref{app:impl}).

\textbf{Effect of the compression tolerance $\tau$.} The tolerance sets how
closely the compressed schedule follows the dense gain $\hat g$. We vary it
over $\tau \in \{0.02, 0.05, 0.1, 0.2\}$. As seen in Table~\ref{tab:inflation}, a coarser tolerance leaves more residual inflation, most in the high-$k$ group, where the gain varies fastest over time, and (2) $\tau = 0.05$ is the coarsest tolerance that keeps the inflation within $10\%$.

\textbf{Effect of the subject mask.} The mask restricts the correction to the
subject. We remove it and apply the gain schedule to every pixel, the
background included. As seen in Table~\ref{tab:ablation}, CLIP-T drops by $0.008$
to $0.282$, below plain CFG's $0.289$: attenuating the background costs prompt
fidelity. The mask is what lets ReGain leave prompt fidelity intact while
correcting the subject.

\textbf{Compute overhead.} Calibration runs once per fine-tuned model and takes $12$ minutes per subject on one RTX A4000 for SD1.5. At sampling, ReGain takes $1.1\times$ the wall-clock time of plain CFG ($5.4$ vs $4.9$ s per image on the same GPU), mostly for
building the mask from the attention maps.

\section{Limitations} 
ReGain corrects the inflation at sampling time, using the base model's guidance as an imperfectly calibrated reference, so it cannot fully match $\Mreal$. A training-time correction during personalization on synthetic images could close this gap. ReGain also relies on a subject mask from existing segmentation methods, so where the mask misses part of the subject or spills onto the background, it attenuates the guidance in the wrong region. In addition, we measure the inflation only at the model's output, in its noise predictions. Mechanistically tracing where it originates inside the network, e.g., in the cross-attention in intermediate activations, is left for future work.

\section{Conclusion}
We studied how personalizing a text-to-image diffusion model on synthetic
rather than real images of a subject affects its outputs. With the base model
and the DreamBooth recipe held fixed, the change of training data alone inflates the CFG term as the angle between the conditional
and unconditional noise predictions widens. This happens most strongly in high frequencies, with its strength varying at each denoising step. Building on these findings, ReGain measures the inflation once against base model (before personalization) and attenuates the guidance per frequency band and step
inside the subject region. It needs no real photographs or retraining, recovers
$51$--$64\%$ of the subject-fidelity gap on SD1.5, improves subject fidelity on
SDXL and SD3.5, while preserving prompt fidelity.

\subsection*{AI use statement}

In this work, we used a generative AI coding assistant to help write measurement
and figure-generation code and to help draft and edit the manuscript text. All
experimental designs, measurements, and claims were specified, executed, and
verified by the authors; AI-assisted code was reviewed and its outputs cross-checked by the authors. We
take responsibility for the final content of this work, including text, claims,
and artifacts produced with the aid of generative AI.

\subsection*{Ethics statement}

Our study uses no human-subject data or personally identifiable information;
subject images come from the public DreamBooth dataset, and we follow the
licenses of Stable Diffusion v1.5 and that dataset.

\subsection*{Reproducibility statement}

All models and data are public (Stable Diffusion v1.5, DreamBooth dataset). Full
implementation details, including the fine-tuning recipe, sampler settings,
guidance weights, and seeds, are given in Appendix~\ref{app:impl}.

\bibliography{iclr2027_conference}

@article{ho2022classifier,
  title={Classifier-Free Diffusion Guidance},
  author={Ho, Jonathan and Salimans, Tim},
  journal={arXiv preprint arXiv:2207.12598},
  year={2022}
}

@inproceedings{rombach2022high,
  title={High-resolution image synthesis with latent diffusion models},
  author={Rombach, Robin and Blattmann, Andreas and Lorenz, Dominik and Esser, Patrick and Ommer, Bj{\"o}rn},
  booktitle={2022 IEEE/CVF conference on computer vision and pattern recognition (CVPR)},
  pages={10674--10685},
  year={2022},
  organization={ieee}
}

@inproceedings{ruiz2023dreambooth,
  title={Dreambooth: Fine tuning text-to-image diffusion models for subject-driven generation},
  author={Ruiz, Nataniel and Li, Yuanzhen and Jampani, Varun and Pritch, Yael and Rubinstein, Michael and Aberman, Kfir},
  booktitle={2023 IEEE/CVF Conference on Computer Vision and Pattern Recognition (CVPR)},
  pages={22500--22510},
  year={2023},
  organization={IEEE}
}

@article{shumailov2024ai,
  title={AI models collapse when trained on recursively generated data},
  author={Shumailov, Ilia and Shumaylov, Zakhar and Zhao, Yiren and Papernot, Nicolas and Anderson, Ross and Gal, Yarin},
  journal={Nature},
  volume={631},
  number={8022},
  pages={755--759},
  year={2024},
  publisher={Nature Publishing Group UK London}
}

@inproceedings{alemohammad2024self,
title={Self-Consuming Generative Models Go {MAD}},
author={Sina Alemohammad and Josue Casco-Rodriguez and Lorenzo Luzi and Ahmed Imtiaz Humayun and Hossein Babaei and Daniel LeJeune and Ali Siahkoohi and Richard Baraniuk},
booktitle={The Twelfth International Conference on Learning Representations},
year={2024},
url={https://openreview.net/forum?id=ShjMHfmPs0}
}

@inproceedings{kynkaanniemi2024applying,
title={Applying Guidance in a Limited Interval Improves Sample and Distribution Quality in Diffusion Models},
author={Tuomas Kynk{\"a}{\"a}nniemi and Miika Aittala and Tero Karras and Samuli Laine and Timo Aila and Jaakko Lehtinen},
booktitle={The Thirty-eighth Annual Conference on Neural Information Processing Systems},
year={2024},
url={https://openreview.net/forum?id=nAIhvNy15T}
}

@inproceedings{karras2024guiding,
  title={Guiding a Diffusion Model with a Bad Version of Itself},
  author={Karras, Tero and Aittala, Miika and Kynk{\"a}{\"a}nniemi, Tuomas and Lehtinen, Jaakko and Aila, Timo and Laine, Samuli},
  booktitle={Advances in Neural Information Processing Systems},
  year={2024}
}

@inproceedings{sadat2025eliminating,
title={Eliminating Oversaturation and Artifacts of High Guidance Scales in Diffusion Models},
author={Seyedmorteza Sadat and Otmar Hilliges and Romann M. Weber},
booktitle={The Thirteenth International Conference on Learning Representations},
year={2025},
url={https://openreview.net/forum?id=e2ONKX6qzJ}
}

@inproceedings{shen2024rethinking,
  title={Rethinking the Spatial Inconsistency in Classifier-Free Diffusion Guidance},
  author={Shen, Dazhong and Song, Guanglu and Xue, Zeyue and Wang, Fu-Yun and Liu, Yu},
  booktitle={Proceedings of the IEEE/CVF Conference on Computer Vision and Pattern Recognition},
  year={2024}
}

@inproceedings{sadat2024cads,
  title={{CADS}: Unleashing the Diversity of Diffusion Models through Condition-Annealed Sampling},
  author={Sadat, Seyedmorteza and Buhmann, Jakob and Bradley, Derek and Hilliges, Otmar and Weber, Romann M.},
  booktitle={International Conference on Learning Representations},
  year={2024}
}

@inproceedings{lin2024common,
  title={Common Diffusion Noise Schedules and Sample Steps are Flawed},
  author={Lin, Shanchuan and Liu, Bingchen and Li, Jiashi and Yang, Xiao},
  booktitle={Proceedings of the IEEE/CVF Winter Conference on Applications of Computer Vision},
  year={2024}
}

@inproceedings{bertrand2024stability,
title={On the Stability of Iterative Retraining of Generative Models on their own Data},
author={Quentin Bertrand and Joey Bose and Alexandre Duplessis and Marco Jiralerspong and Gauthier Gidel},
booktitle={The Twelfth International Conference on Learning Representations},
year={2024},
url={https://openreview.net/forum?id=JORAfH2xFd}
}

@inproceedings{kumari2023multi,
  author = {Kumari, Nupur and Zhang, Bingliang and Zhang, Richard and Shechtman, Eli and Zhu, Jun-Yan},
  title = {Multi-Concept Customization of Text-to-Image Diffusion},
  booktitle = {CVPR},
  year = {2023},
}

@inproceedings{hu2022lora,
  title={Lo{RA}: Low-Rank Adaptation of Large Language Models},
  author={Hu, Edward J. and Shen, Yelong and Wallis, Phillip and Allen-Zhu, Zeyuan and Li, Yuanzhi and Wang, Shean and Wang, Lu and Chen, Weizhu},
  booktitle={International Conference on Learning Representations},
  year={2022}
}

@inproceedings{shah2024ziplora,
  title={Zip{L}o{RA}: Any Subject in Any Style by Effectively Merging {L}o{RA}s},
  author={Shah, Viraj and Ruiz, Nataniel and Cole, Forrester and Lu, Erika and Lazebnik, Svetlana and Li, Yuanzhen and Jampani, Varun},
  booktitle={European Conference on Computer Vision},
  year={2024}
}

@inproceedings{gerstgrasser2024accumulate,
title={Is Model Collapse Inevitable? Breaking the Curse of Recursion by Accumulating Real and Synthetic Data},
author={Matthias Gerstgrasser and Rylan Schaeffer and Apratim Dey and Rafael Rafailov and Tomasz Korbak and Henry Sleight and Rajashree Agrawal and John Hughes and Dhruv Bhandarkar Pai and Andrey Gromov and Dan Roberts and Diyi Yang and David L. Donoho and Sanmi Koyejo},
booktitle={First Conference on Language Modeling},
year={2024},
url={https://openreview.net/forum?id=5B2K4LRgmz}
}

@inproceedings{
yoon2025model,
title={Model Collapse in the Self-Consuming Chain of Diffusion Finetuning: A Novel Perspective from Quantitative Trait Modeling},
author={Youngseok Yoon and Dainong Hu and Iain Weissburg and Yao Qin and Haewon Jeong},
booktitle={ICLR 2025 Workshop on Navigating and Addressing Data Problems for Foundation Models},
year={2025},
url={https://openreview.net/forum?id=1MIgdKsvjX}
}

@inproceedings{alemohammad2025self,
title={Self-Improving Diffusion Models With Synthetic Data},
author={Sina Alemohammad and Ahmed Imtiaz Humayun and Shruti Agarwal and John Collomosse and Richard Baraniuk},
booktitle={Scaling Self-Improving Foundation Models without Human Supervision},
year={2025},
url={https://openreview.net/forum?id=FHTCV0iE06}
}

@inproceedings{park2025steering,
  title={Steering guidance for personalized text-to-image diffusion models},
  author={Park, Sunghyun and Choi, Seokeon and Park, Hyoungwoo and Yun, Sungrack},
  booktitle={2025 IEEE/CVF International Conference on Computer Vision (ICCV)},
  pages={15907--15916},
  year={2025},
  organization={IEEE}
}

@inproceedings{jeong2025mindiff,
  title={MINDiff: Mask-Integrated Negative Attention for Controlling Overfitting in Text-to-Image Personalization},
  author={Jeong, Seulgi and Kim, Jaeil},
  booktitle={2025 IEEE/CVF International Conference on Computer Vision Workshops (ICCVW)},
  pages={6981--6990},
  year={2025},
  organization={IEEE}
}

@article{ye2023ip,
  title={Ip-adapter: Text compatible image prompt adapter for text-to-image diffusion models},
  author={Ye, Hu and Zhang, Jun and Liu, Sibo and Han, Xiao and Yang, Wei},
  journal={arXiv preprint arXiv:2308.06721},
  year={2023}
}

@inproceedings{tan2025ominicontrol,
  title={Ominicontrol: Minimal and universal control for diffusion transformer},
  author={Tan, Zhenxiong and Liu, Songhua and Yang, Xingyi and Xue, Qiaochu and Wang, Xinchao},
  booktitle={2025 IEEE/CVF International Conference on Computer Vision (ICCV)},
  pages={14940--14950},
  year={2025},
  organization={IEEE}
}

@inproceedings{wu2025less,
  title={Less-to-more generalization: Unlocking more controllability by in-context generation},
  author={Wu, Shaojin and Huang, Mengqi and Wu, Wenxu and Cheng, Yufeng and Ding, Fei and He, Qian},
  booktitle={2025 IEEE/CVF International Conference on Computer Vision (ICCV)},
  pages={18682--18692},
  year={2025},
  organization={IEEE}
}

@inproceedings{gal2023image,
title={An Image is Worth One Word: Personalizing Text-to-Image Generation using Textual Inversion},
author={Rinon Gal and Yuval Alaluf and Yuval Atzmon and Or Patashnik and Amit Haim Bermano and Gal Chechik and Daniel Cohen-Or},
booktitle={The Eleventh International Conference on Learning Representations },
year={2023},
url={https://openreview.net/forum?id=NAQvF08TcyG}
}

@article{lee2024direct,
  title={Direct consistency optimization for robust customization of text-to-image diffusion models},
  author={Lee, Kyungmin and Kwak, Sangkyung and Sohn, Kihyuk and Shin, Jinwoo},
  journal={Advances in neural information processing systems},
  volume={37},
  pages={103269--103304},
  year={2024}
}

@inproceedings{kim2026preserve,
  title={Preserve and Personalize: Personalized Text-to-Image Diffusion Models without Distributional Drift},
  author={Kim, Gihoon and Park, Hyungjin and Kim, Taesup},
  booktitle={International Conference on Learning Representations},
  volume={2026},
  pages={15591--15615},
  year={2026}
}

@inproceedings{kumari2025generating,
  title={Generating multi-image synthetic data for text-to-image customization},
  author={Kumari, Nupur and Yin, Xi and Zhu, Jun-Yan and Misra, Ishan and Azadi, Samaneh},
  booktitle={2025 IEEE/CVF International Conference on Computer Vision (ICCV)},
  pages={16524--16534},
  year={2025},
  organization={IEEE}
}

@inproceedings{li2025iccustom,
title={{IC}-Custom: Diverse Image Customization via In-Context Learning},
author={Yaowei Li and Xiaoyu Li and Zhaoyang Zhang and Yuxuan Bian and Gan Liu and Xinyuan Li and Jiale Xu and Wenbo Hu and yating liu and Lingen Li and Jing Cai and Yuexian Zou and Yancheng He and Ying Shan},
booktitle={The Fourteenth International Conference on Learning Representations},
year={2026},
url={https://openreview.net/forum?id=gv2cr8kABL}
}

@inproceedings{song2021ddim,
title={Denoising Diffusion Implicit Models},
author={Jiaming Song and Chenlin Meng and Stefano Ermon},
booktitle={International Conference on Learning Representations},
year={2021},
url={https://openreview.net/forum?id=St1giarCHLP}
}

@article{oquab2024dinov2,
title={{DINO}v2: Learning Robust Visual Features without Supervision},
author={Maxime Oquab and Timoth{\'e}e Darcet and Th{\'e}o Moutakanni and Huy V. Vo and Marc Szafraniec and Vasil Khalidov and Pierre Fernandez and Daniel HAZIZA and Francisco Massa and Alaaeldin El-Nouby and Mido Assran and Nicolas Ballas and Wojciech Galuba and Russell Howes and Po-Yao Huang and Shang-Wen Li and Ishan Misra and Michael Rabbat and Vasu Sharma and Gabriel Synnaeve and Hu Xu and Herve Jegou and Julien Mairal and Patrick Labatut and Armand Joulin and Piotr Bojanowski},
journal={Transactions on Machine Learning Research},
issn={2835-8856},
year={2024},
url={https://openreview.net/forum?id=a68SUt6zFt},
note={Featured Certification}
}

@inproceedings{avrahami2024chosen,
 title={The chosen one: Consistent characters in text-to-image diffusion models},
  author={Avrahami, Omri and Hertz, Amir and Vinker, Yael and Arar, Moab and Fruchter, Shlomi and Fried, Ohad and Cohen-Or, Daniel and Lischinski, Dani},
  booktitle={ACM SIGGRAPH 2024 conference papers},
  pages={1--12},
  year={2024}
}

@inproceedings{podell2024sdxl,
title={{SDXL}: Improving Latent Diffusion Models for High-Resolution Image Synthesis},
author={Dustin Podell and Zion English and Kyle Lacey and Andreas Blattmann and Tim Dockhorn and Jonas M{\"u}ller and Joe Penna and Robin Rombach},
booktitle={The Twelfth International Conference on Learning Representations},
year={2024},
url={https://openreview.net/forum?id=di52zR8xgf}
}

@inproceedings{esser2024scaling,
  title={Scaling rectified flow transformers for high-resolution image synthesis},
  author={Esser, Patrick and Kulal, Sumith and Blattmann, Andreas and Entezari, Rahim and M{\"u}ller, Jonas and Saini, Harry and Levi, Yam and Lorenz, Dominik and Sauer, Axel and Boesel, Frederic and others},
  booktitle={Forty-first international conference on machine learning},
  year={2024}
}

@inproceedings{radford2021learning,
  title={Learning transferable visual models from natural language supervision},
  author={Radford, Alec and Kim, Jong Wook and Hallacy, Chris and Ramesh, Aditya and Goh, Gabriel and Agarwal, Sandhini and Sastry, Girish and Askell, Amanda and Mishkin, Pamela and Clark, Jack and others},
  booktitle={International conference on machine learning},
  pages={8748--8763},
  year={2021},
  organization={PMLR}
}

@inproceedings{caron2021emerging,
  title={Emerging properties in self-supervised vision transformers},
  author={Caron, Mathilde and Touvron, Hugo and Misra, Ishan and J{\'e}gou, Herv{\'e} and Mairal, Julien and Bojanowski, Piotr and Joulin, Armand},
  booktitle={2021 IEEE/CVF international conference on computer vision (ICCV)},
  pages={9630--9640},
  year={2021},
  organization={IEEE}
}

@inproceedings{kim2025seg4diff,
title={Seg4Diff: Unveiling Open-Vocabulary Semantic Segmentation in Text-to-Image Diffusion Transformers},
author={Chaehyun Kim and Heeseong Shin and Eunbeen Hong and Heeji Yoon and Anurag Arnab and Paul Hongsuck Seo and Sunghwan Hong and Seungryong Kim},
booktitle={The Thirty-ninth Annual Conference on Neural Information Processing Systems},
year={2025},
url={https://openreview.net/forum?id=ENp2kCdYE8}
}

@inproceedings{chan2024subjectagnostic,
  title={Improving subject-driven image synthesis with subject-agnostic guidance},
  author={Chan, Kelvin CK and Zhao, Yang and Jia, Xuhui and Yang, Ming-Hsuan and Wang, Huisheng},
  booktitle={2024 IEEE/CVF Conference on Computer Vision and Pattern Recognition (CVPR)},
  pages={6733--6742},
  year={2024},
  organization={IEEE}
}

@article{sadat2025fdg,
  title={Guidance in the frequency domain enables high-fidelity sampling at low cfg scales},
  author={Sadat, Seyedmorteza and Vontobel, Tobias and Salehi, Farnood and Weber, Romann M},
  journal={arXiv preprint arXiv:2506.19713},
  year={2025}
}

@inproceedings{rahaman2019spectral,
  title={On the spectral bias of neural networks},
  author={Rahaman, Nasim and Baratin, Aristide and Arpit, Devansh and Draxler, Felix and Lin, Min and Hamprecht, Fred and Bengio, Yoshua and Courville, Aaron},
  booktitle={International conference on machine learning},
  pages={5301--5310},
  year={2019},
  organization={PMLR}
}

@inproceedings{chung2025cfgpp,
title={{CFG}++: Manifold-constrained Classifier Free Guidance for Diffusion Models},
author={Hyungjin Chung and Jeongsol Kim and Geon Yeong Park and Hyelin Nam and Jong Chul Ye},
booktitle={The Thirteenth International Conference on Learning Representations},
year={2025},
url={https://openreview.net/forum?id=E77uvbOTtp}
}
\bibliographystyle{iclr2027_conference}

\clearpage
\appendix

\section*{Appendix}

\etocdepthtag.toc{mtappendix}
\etocsettagdepth{mtmain}{none}
\etocsettagdepth{mtappendix}{subsection}
\etocsettocstyle{}{}
\tableofcontents
\vspace{1em}

\section{Synthetic training images as the source of the $\Delta$ inflation}
\label{app:synthetic-source}

Section~\ref{sec:findings} attributes the $\Delta$ inflation of $\Msyn$ to its
training images having been generated by $\Mreal$. Two other properties of the
generated images could account for it: they could be less diverse than the real
photographs $\Mreal$ was trained on, or they could inherit the guidance weight at
which $\Mreal$ sampled them. Neither accounts for the inflation. Nor is the
inflation specific to fine-tuning without the prior-preservation loss of
DreamBooth.

\subsection{Diversity of the training images}
\label{app:diversity}

$\Mreal$ and $\Msyn$ are fine-tuned on different images, the subject's real
photographs and the five images that $\Mreal$ generated. The $\Delta$ inflation of
Section~\ref{sec:findings} could therefore be attributed to the synthetic set
being less diverse than the real one rather than to its origin. For each set we
measure diversity as the mean, over all pairs of images in the set, of one minus
the cosine similarity between their CLIP ViT-B/32 embeddings
(Section~\ref{sec:setup}), which does not depend on the number of images, and
report the mean and standard deviation over the $30$ subjects:
$0.142 \pm 0.050$ for the real photographs and $0.140 \pm 0.052$ for the five
images that $\Mreal$ generated. The two sets are comparably diverse, so what
separates $\Mreal$ from $\Msyn$ is the origin of their training images and not their
diversity.

\subsection{Guidance weight of the training images}
\label{app:gen-guidance}

Every $\Msyn$ in the paper is trained on images that $\Mreal$ sampled at the
guidance weight of Table~\ref{tab:impl}, $w = 7.5$ on Stable Diffusion v1.5. A
higher guidance weight drives each sample further along the guidance direction,
so the inflation could be inherited from the weight at which the training
images were sampled and would then fall at lower weights. For dog6, the subject
of Figure~\ref{fig:regain}, we resample the five training images from the same
$\Mreal$ at $w = 5$ and $w = 3$ with the same prompts and seeds, train $\Msyn$ on
each set with the Stable Diffusion v1.5 DreamBooth recipe of
Table~\ref{tab:impl}, unchanged, and repeat the measurement of
Table~\ref{tab:inflation}. Table~\ref{tab:gen-guidance} reports the inflation
at the calibration states, before correction, by band group; the value at
$7.5$, $5.57$, sits within the $30$-subject mean of Table~\ref{tab:inflation},
$5.31 \pm 1.26$. The inflation stays large at all three weights and keeps its
ordering from the low-$k$ to the high-$k$ group. Lowering the weight from $7.5$
to $3$ reduces it by about $12\%$, and most of that drop is already reached at
$5$.

\begin{table}[h]
\centering
\caption{$\Delta$ inflation of dog6 against the base model, by band group, for
$\Msyn$ trained on images sampled from $\Mreal$ at three guidance weights (root mean
square over the group's bands and all steps at the calibration states, as the
Before column of Table~\ref{tab:inflation}). The $7.5$ row is the $\Msyn$ used
throughout the paper.}
\label{tab:gen-guidance}
\begin{tabular}{lcccc}
\toprule
Guidance weight $w$ & Low $k$ & Mid $k$ & High $k$ & All \\
\midrule
7.5 & 4.16 & 4.71 & 6.55 & 5.57 \\
5.0 & 3.86 & 4.28 & 5.80 & 4.98 \\
3.0 & 3.50 & 4.09 & 5.79 & 4.88 \\
\bottomrule
\end{tabular}
\end{table}

\subsection{Prior-preservation loss}
\label{app:prior-pres}

DreamBooth \citep{ruiz2023dreambooth} can add a prior-preservation loss, which
also trains the model on images of the subject's class that the base model
generates under the class prompt, so that the class noun does not drift toward
the subject. The models in the paper are fine-tuned without it
(Table~\ref{tab:impl}). To test whether it would prevent the inflation, for
dog6 we fine-tune $\Mreal$ and $\Msyn$ as in the main experiments with this
loss added, training for $800$ steps following the Diffusers recommendation,
with $\Msyn$ again trained on five images generated by $\Mreal$, and repeat the
measurement of Table~\ref{tab:inflation}. At the calibration states, before
correction and computed as in Table~\ref{tab:gen-guidance}, the inflation
against the base model is $7.38$, $7.35$ and $9.47$ in the low-, mid- and
high-$k$ groups and $8.41$ over all bands. The prior-preservation loss
therefore does not remove the inflation, which remains largest in the high-$k$
group. Since it leaves the inflation in place, we keep the simpler setup
without it.

\subsection{The $\Delta$ inflation on all subjects}
\label{app:all-subjects}
\begin{figure}[h]
\centering
\includegraphics[width=\linewidth]{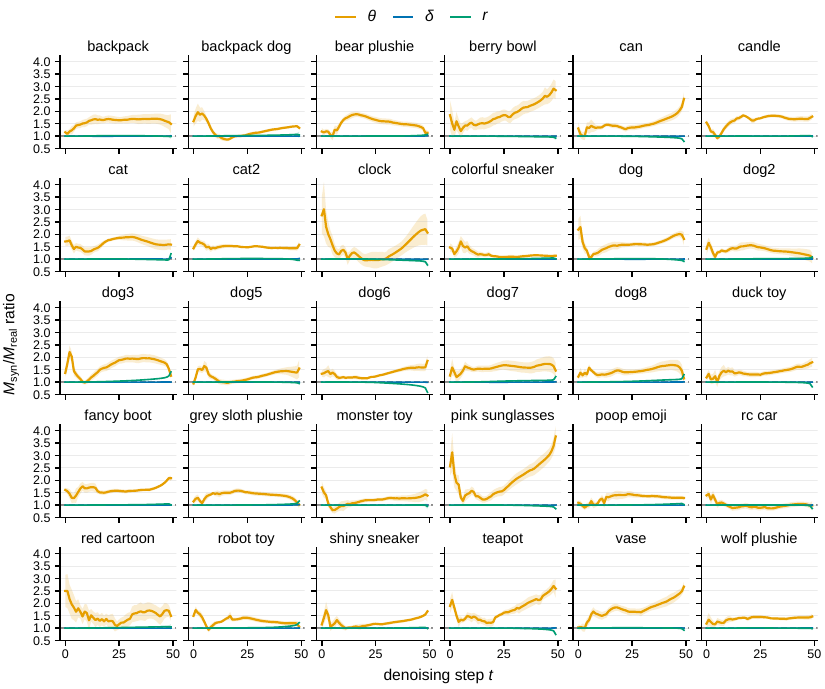}
\caption{Per-step $\Msyn/\Mreal$ ratio of $\theta$, $\delta$ and $r$ for the
$30$ DreamBooth subjects, as in Figure~\ref{fig:findings}a. Mean over $5$
seeds, shading is the standard error.}
\label{fig:app-all-subjects}
\end{figure}
We repeat the measurement of finding~1 on all $30$ DreamBooth subjects, each
with its own $\Mreal$ and $\Msyn$, under the subject prompt with the subject's
class noun and $5$ seeds. Figure~\ref{fig:app-all-subjects} shows the per-step
ratios of $\theta$, $\delta$ and $r$ for every subject. Averaged over the $50$
steps, $\|\Delta\|$ of $\Msyn$ exceeds that of $\Mreal$ in $29$ of the $30$
subjects, by $48 \pm 22\%$ (mean and standard deviation over subjects), and
$\theta$ by $48 \pm 22\%$, while $\delta$ and $r$ change by $-0.1 \pm 0.1\%$
and $0.6 \pm 2.2\%$. The exception, rc\_car, shows no excess in either.

\section{Subject mask}
\label{app:mask}

The subject mask $m$ of \eqref{eq:masked-update} is built from existing segmentation methods, used unchanged.

\textbf{Saliency maps.} On Stable Diffusion v1.5 and SDXL we use the segmentation
stage of S-CFG \citep{shen2024rethinking} unchanged: the cross-attention maps of
the conditional pass at the two coarsest U-Net resolutions ($16\times16$ and
$8\times8$ on Stable Diffusion v1.5) are refined by propagation over the
self-attention affinity graph (SSGC, four hops), normalized per token to unit
spatial mean, averaged over layers, upsampled to the coarsest resolution and
smoothed with a $3\times3$ Gaussian ($\sigma = 0.5$). This gives one saliency
map $s_j$ per prompt token. Stable Diffusion 3.5 has no cross-attention. There we
follow Seg4Diff \citep{kim2025seg4diff} and read the maps from the joint
attention of transformer block $9$, taking the image queries against the $77$
CLIP text keys, with the T5 encoder omitted from this one evaluation, on the
$64\times64$ token grid and without smoothing, at the cost of one additional
conditional evaluation per sampling step.

\begin{figure}[b!]
\centering
\includegraphics[width=\linewidth]{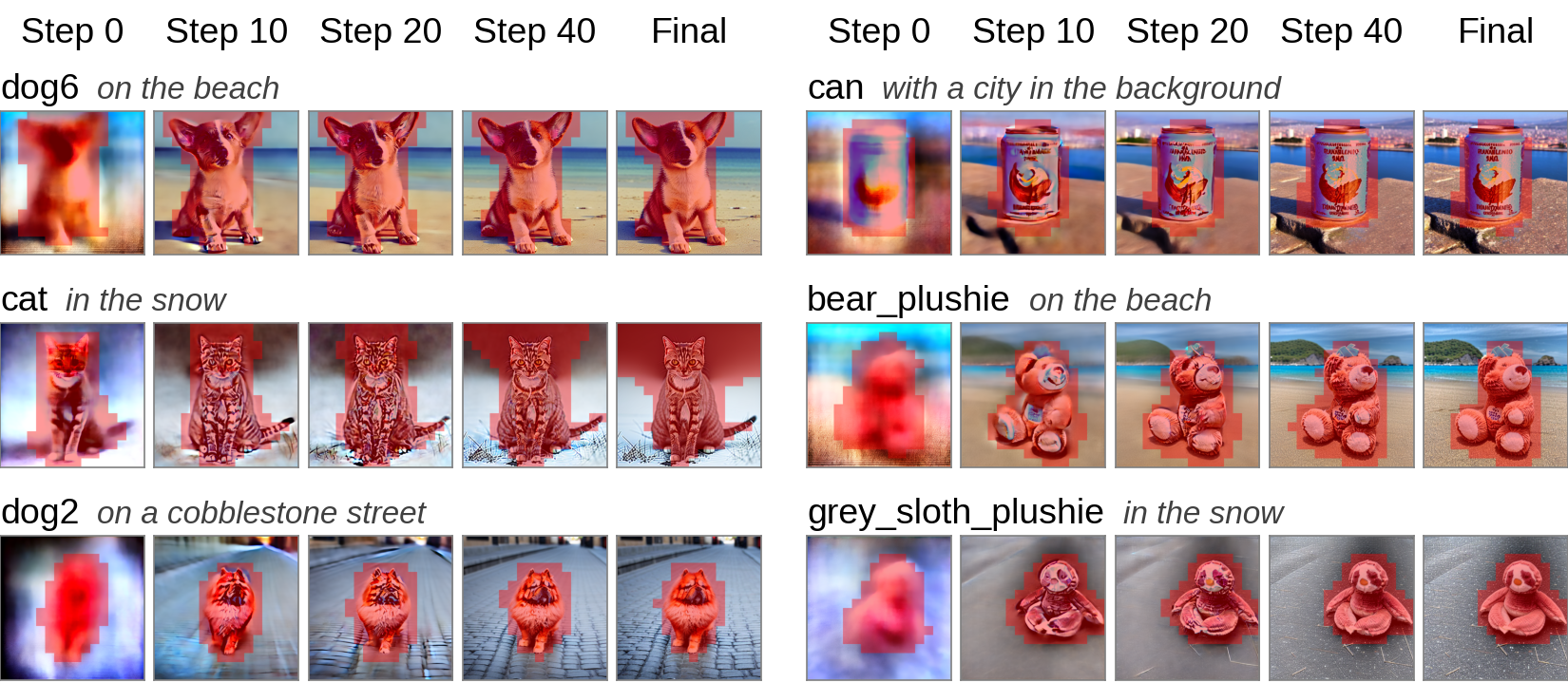}
\caption{\textbf{Subject mask $m$ during sampling.} The mask of
\eqref{eq:fgbg-mask} (red) at steps $0$, $10$, $20$ and $40$ of a $50$-step
plain-CFG trajectory of $\Msyn$, drawn on the clean image predicted at that
step, and the mask of the last step on the final sample. Stable Diffusion v1.5,
DreamBooth, seed $100$. Each row gives the subject and the scene phrase of its
evaluation prompt \emph{``a [V] $\langle$class$\rangle$ \ldots''}
(Appendix~\ref{app:eval-prompts}).}
\label{fig:app-mask}
\end{figure}

\textbf{Competition rule.} The mask is a per-pixel competition between two token
sets with no free parameters:
\begin{equation}
m(p) \;=\; \mathbb{1}\!\left[\; \max_{j \in \mathcal{S}} s_j(p)
\;>\; \max_{j \in \mathcal{B}} s_j(p) \;\right],
\label{eq:fgbg-mask}
\end{equation}
where $\mathcal{S}$ holds the sub-word tokens of the identifier and the class
noun and $\mathcal{B}$ the prompt's scene content words. Function words and
generic framing words such as \emph{photo} are excluded through a fixed stopword
list: after per-token normalization their maps are nearly flat at unit height
and would outcompete the subject wherever its saliency is not sharply peaked,
collapsing the mask to the subject's most discriminative parts. The maximum
within each set keeps the rule invariant to the number of words in it. Pixels
claimed by neither set resolve to the subject, so the mask errs toward
over-coverage of texture-free background. Such pixels receive the rescaled
guidance term of \eqref{eq:masked-update}. The same construction is used
at every sampling step, across seeds, prompts and subjects
(Figure~\ref{fig:app-mask}).

\textbf{Mask at calibration.} The calibration prompts of
Section~\ref{sec:regain} name no scene content, so $\mathcal{B}$ would be
empty. The mask is therefore read from one auxiliary evaluation of $\Msyn$ at the
same forward-noised state under the subject prompt extended with the fixed
phrase ``in a scene'', whose single content word fills $\mathcal{B}$. The rule
\eqref{eq:fgbg-mask} is otherwise unchanged and no caption of the training
images is used. Where the captions are known they permit a check: masks built with the generic phrase agree with masks built from each training image's own caption at IoU $0.86$, averaged over the five training images and all $50$ steps. At sampling, the same auxiliary evaluation supplies the mask when the sampled prompt names no scene content.

\section{Implementation details}
\label{app:impl}

Table~\ref{tab:impl} lists the fine-tuning and sampling parameters of the four
settings of Table~\ref{tab:metrics}. Within a setting, $\Mreal$ and $\Msyn$ are
trained with the same procedure and hyperparameters and differ only in their
training images: $\Mreal$ is trained on the subject's real photographs and $\Msyn$ on
the five images that $\Mreal$ generated. The identifier [V] is the token
\emph{monadikos} and the fine-tuning prompt is \emph{``a photo of monadikos
$\langle$class$\rangle$''} throughout.

\begin{table}[h]
\centering
\caption{Fine-tuning and sampling settings. One column per setting of
Table~\ref{tab:metrics}. LoRA adapters use a scaling factor $\alpha$ equal to the
rank and are merged into the base model weights before sampling. All text
encoders are frozen and no prior-preservation loss is used. The synthetic
training set is sampled with each base model's default scheduler and evaluation
uses DDIM, except on Stable Diffusion 3.5, which uses its flow-matching Euler
scheduler (shift $3.0$) in both cases.}
\label{tab:impl}
\small
\setlength{\tabcolsep}{4pt}
\begin{tabular}{lcccc}
\toprule
 & SD v1.5 & SD v1.5 & SDXL 1.0 & SD 3.5 Medium \\
 & DreamBooth & DreamBooth-LoRA & DreamBooth-LoRA & DreamBooth-LoRA \\
\midrule
\multicolumn{5}{l}{\emph{Fine-tuning}} \\
Trained weights & full U-Net & U-Net attention & U-Net attention & MMDiT \\
LoRA rank & -- & 16 & 4 & 4 \\
Learning rate (constant) & $5\times10^{-6}$ & $10^{-4}$ & $5\times10^{-4}$ & $4\times10^{-4}$ \\
Steps & 500 & 500 & 500 & 600 \\
Batch size & 1 & 1 & 1 & 1 \\
Gradient accumulation & 1 & 1 & 1 & 4 \\
Resolution & 512 & 512 & 1024 & 512 \\
Precision & fp32 & fp32 & fp16 & bf16 \\
Subjects & 30 & 30 & 30 & 30 \\
\midrule
\multicolumn{5}{l}{\emph{Synthetic training set of $\Msyn$ (generated by $\Mreal$)}} \\
Images & 5 & 5 & 5 & 5 \\
Sampler, steps & PNDM, 50 & PNDM, 50 & Euler, 50 & flow Euler, 40 \\
Guidance weight & 7.5 & 7.5 & 7.5 & 7.0 \\
Resolution & 512 & 512 & 1024 & 1024 \\
\midrule
\multicolumn{5}{l}{\emph{Evaluation sampling (all models)}} \\
Sampler, steps & DDIM, 50 & DDIM, 50 & DDIM, 50 & flow Euler, 40 \\
Guidance weight $w$ & 7.5 & 7.5 & 7.5 & 7.0 \\
Resolution & 512 & 512 & 1024 & 1024 \\
Seeds & 100--102 & 100--102 & 100--102 & 100--102 \\
\bottomrule
\end{tabular}
\end{table}

\textbf{Synthetic training set.} $\Mreal$ generates one image under each of five
prompt templates shared by all subjects and disjoint from the evaluation prompts:
a close-up photo of the subject, and \emph{``a photo of monadikos
$\langle$class$\rangle$''} completed by one of \emph{outdoors in a backyard on a
sunny day}, \emph{on a couch}, \emph{on a staircase} and \emph{in front of a
brick wall}.

\textbf{Guidance baselines.} The three baselines of Table~\ref{tab:metrics} are
sampled from the same $\Msyn$ on both Stable Diffusion v1.5 settings, with DDIM
for $50$ steps at the seeds and prompts of ReGain and with the settings of their
official implementations. S-CFG \citep{shen2024rethinking} uses $w = 7.5$ and
rescales the guidance of each attention region at every step, with the rate
clipped to $[0.8, 3.0]$. CFG++ \citep{chung2025cfgpp} uses $\lambda = 0.6$, the
value its authors match to $w = 7.5$ at $50$ steps. FDG \citep{sadat2025fdg}
splits the guidance with a one-level Laplacian pyramid and uses $w = 7.5$ on the
high band and $w = 3$ on the low band, the setting its authors report for Stable
Diffusion 2.1.

\textbf{Trajectory measurements.} The measurements of Section~\ref{sec:findings}
are made on one dog subject (dog6), with $\Mreal$ and $\Msyn$ obtained by DreamBooth
fine-tuning of Stable Diffusion v1.5. Both models are sampled with DDIM for $50$
steps at $w = 7.5$, with $10$ seeds shared by both models and one image per
prompt and seed. For the band-resolved comparison, $\Mreal$ is evaluated at the
states that $\Msyn$ visits. The frequency bands are computed on the
$64 \times 64$ latent, which gives $K = 46$ bands, and the low, mid and high
band groups are $k = 1$ to $8$, $9$ to $24$ and $25$ to $45$.

\begin{figure}[!ht]
\centering
\includegraphics[width=\linewidth]{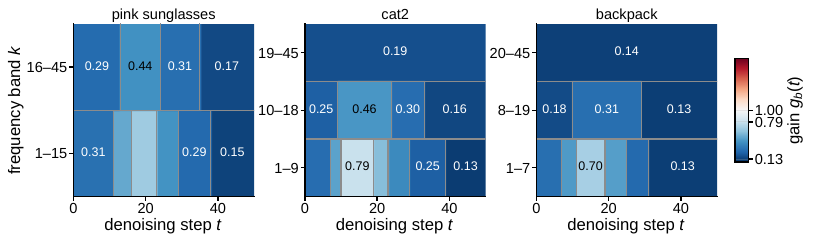}
\caption{\textbf{Estimated gain schedules of three further subjects.} The
compressed schedule $g_b(t)$ at $\tau = 0.05$ of the pink sunglasses, cat2 and
backpack subjects of Figure~\ref{fig:results-single}, Stable Diffusion v1.5
DreamBooth, in the layout of Figure~\ref{fig:regain}: one row per band group,
one column per sampling step, and color and label giving the gain.}
\label{fig:app-schedules}
\end{figure}

\textbf{Schedule estimation.} The estimator \eqref{eq:estimator} is evaluated at
the five training images, each forward-noised with $10$ independent noise draws
at every step, the same draws for both models. The estimated gain $\hat g(k,t)$
has one band per ring of the model's latent grid, $46$ on the $64 \times 64$
latent of Stable Diffusion v1.5 and $91$ on the $128 \times 128$ latent of SDXL
and Stable Diffusion 3.5, and one column per sampling step ($50$, or $40$ on Stable
Diffusion 3.5). The compression uses the root-mean-square tolerance
$\tau = 0.05$ at each of its two stages. The number of cells is determined by
the compression and varies by subject: dog6 has $3$ band groups and $15$ cells
(Figure~\ref{fig:regain}), and the mean over the $30$ subjects of
Table~\ref{tab:inflation} is $18$ cells. The DC band is kept as measured.
At the calibration states, the residual inflation after the compressed schedule
(Table~\ref{tab:inflation}) sits in the last five steps, where the schedule's
time cells are coarsest.
Figure~\ref{fig:app-schedules} shows the schedules of three further subjects.
Across all $30$ subjects, the highest band group is attenuated at every step,
with gains between $0.11$ and $0.79$ (median $0.21$).

\subsection{Evaluation prompts}
\label{app:eval-prompts}

Evaluation uses the official DreamBooth prompt lists \citep{ruiz2023dreambooth}
verbatim, instantiated as \emph{``a monadikos $\langle$class$\rangle$ \ldots''}
with each subject's official class noun. Object subjects use $25$ prompts:
\emph{in the jungle, in the snow, on the beach, on a cobblestone street, on top
of pink fabric, on top of a wooden floor, with a city in the background, with a
mountain in the background, with a blue house in the background, on top of a
purple rug in a forest, with a wheat field in the background, with a tree and
autumn leaves in the background, with the Eiffel Tower in the background,
floating on top of water, floating in an ocean of milk, on top of green grass
with sunflowers around it, on top of a mirror, on top of the sidewalk in a
crowded street, on top of a dirt road, on top of a white rug}, and the property
modifications \emph{red}, \emph{purple}, \emph{shiny}, \emph{wet} and
\emph{cube shaped}. Live subjects use $25$ prompts: the first ten
recontextualization prompts above, the accessorization prompts \emph{wearing a
red hat, wearing a santa hat, wearing a rainbow scarf, wearing a black top hat
and a monocle, in a chef outfit, in a firefighter outfit, in a police outfit,
wearing pink glasses, wearing a yellow shirt, in a purple wizard outfit}, and
the same five property modifications.

\section{Over-guidance artifact metrics}
\label{app:artifact-metrics}

\textbf{Definitions.} Saturation is the mean of the HSV $S$ channel, and
root-mean-square contrast is the standard deviation of ITU-R 601 grayscale
intensity, both on $[0,1]$. The high-band fraction is the share of Fourier
power, with the zero-frequency term removed, beyond a radial cutoff at a
quarter of the Nyquist frequency, the pixel-space image of the low-band
boundary of Section~\ref{sec:findings}. Removing the mean matters: with the DC
term left in the denominator it dominates the total power and scales with
brightness and variance, so the statistic would then track contrast as well as
spectral shape.

\textbf{Measurement region.} The region of Table~\ref{tab:artifacts} is the
mask that steered $\Msyn$, taken once per subject, prompt and seed and applied
to all three methods so that they are compared over identical pixels. Over the
evaluation set it covers $43\%$ of the frame. The high-band fraction needs a
rectangular grid and is measured on the mask's bounding box.

\section{Additional qualitative results}
\label{app:more-results}

\begin{figure}[p]
\centering
\includegraphics[width=0.7\linewidth]{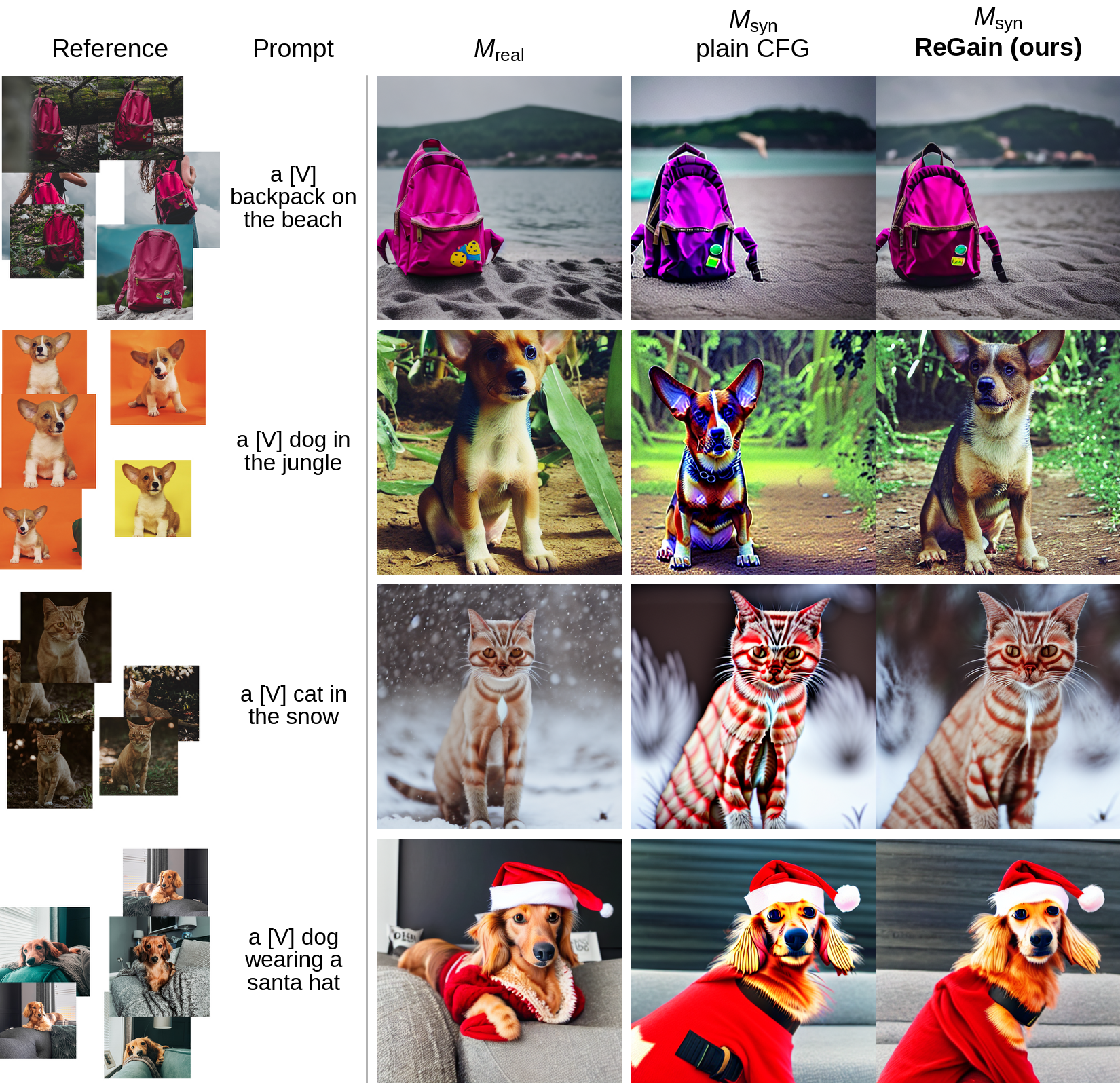}
\caption{\textbf{DreamBooth-LoRA, Stable Diffusion v1.5.} The block of
Figure~\ref{fig:results-single} in the DreamBooth-LoRA setting: the subject's
reference photos, the prompt, and then $\Mreal$, $\Msyn$ under plain CFG, and the
same $\Msyn$ under its estimated gain schedule (ReGain, ours), at one
seed and $w{=}7.5$.
}
\label{fig:app-lora-qual}
\end{figure}

\begin{figure}[p]
\centering
\includegraphics[width=0.7\linewidth]{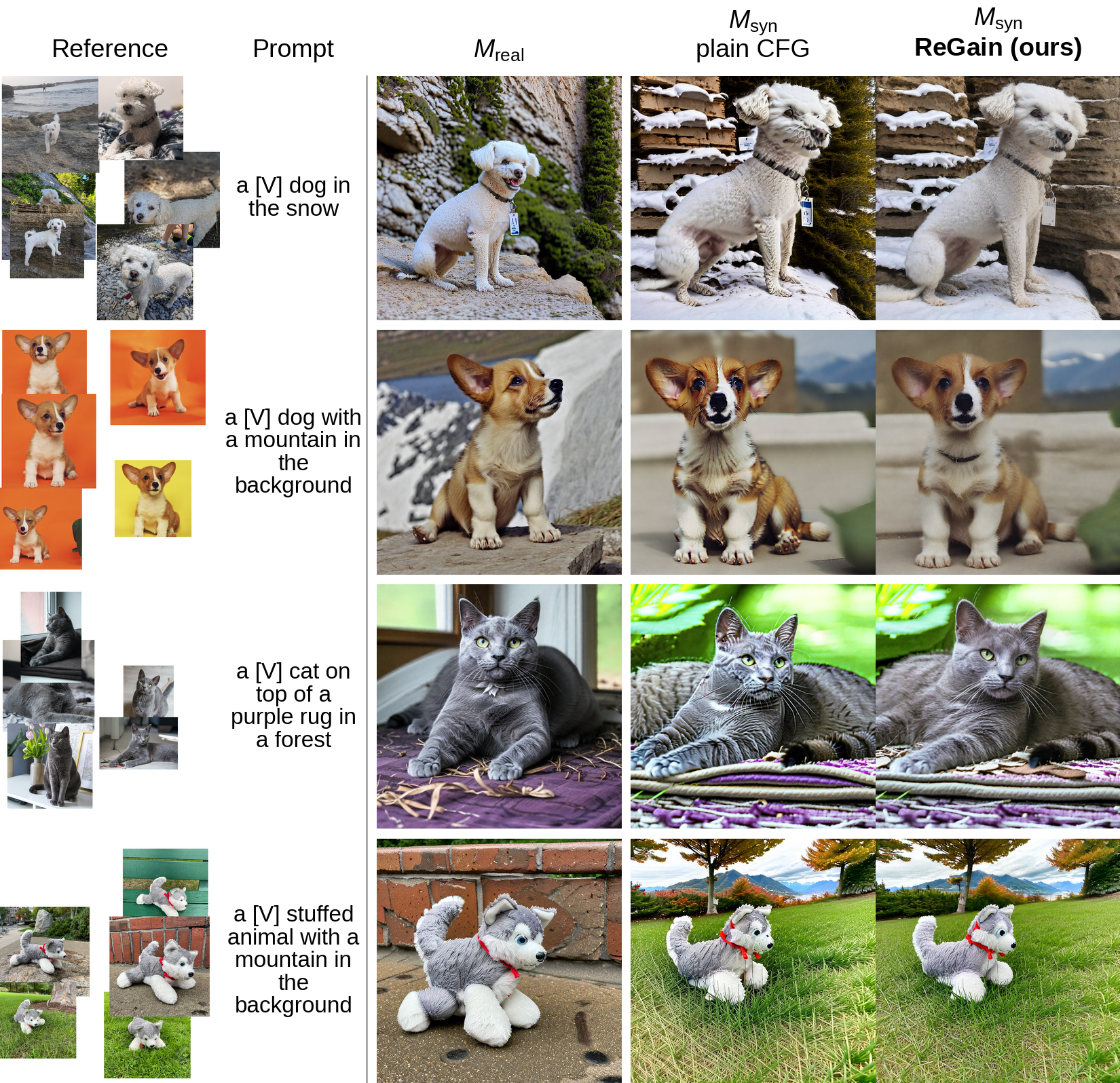}
\caption{\textbf{DreamBooth-LoRA, SDXL base 1.0.} The same block on SDXL at
$1024\times1024$. Columns and protocol follow Figure~\ref{fig:app-lora-qual}.
}
\label{fig:app-sdxl-qual}
\end{figure}

\begin{figure}[p]
\centering
\includegraphics[width=0.7\linewidth]{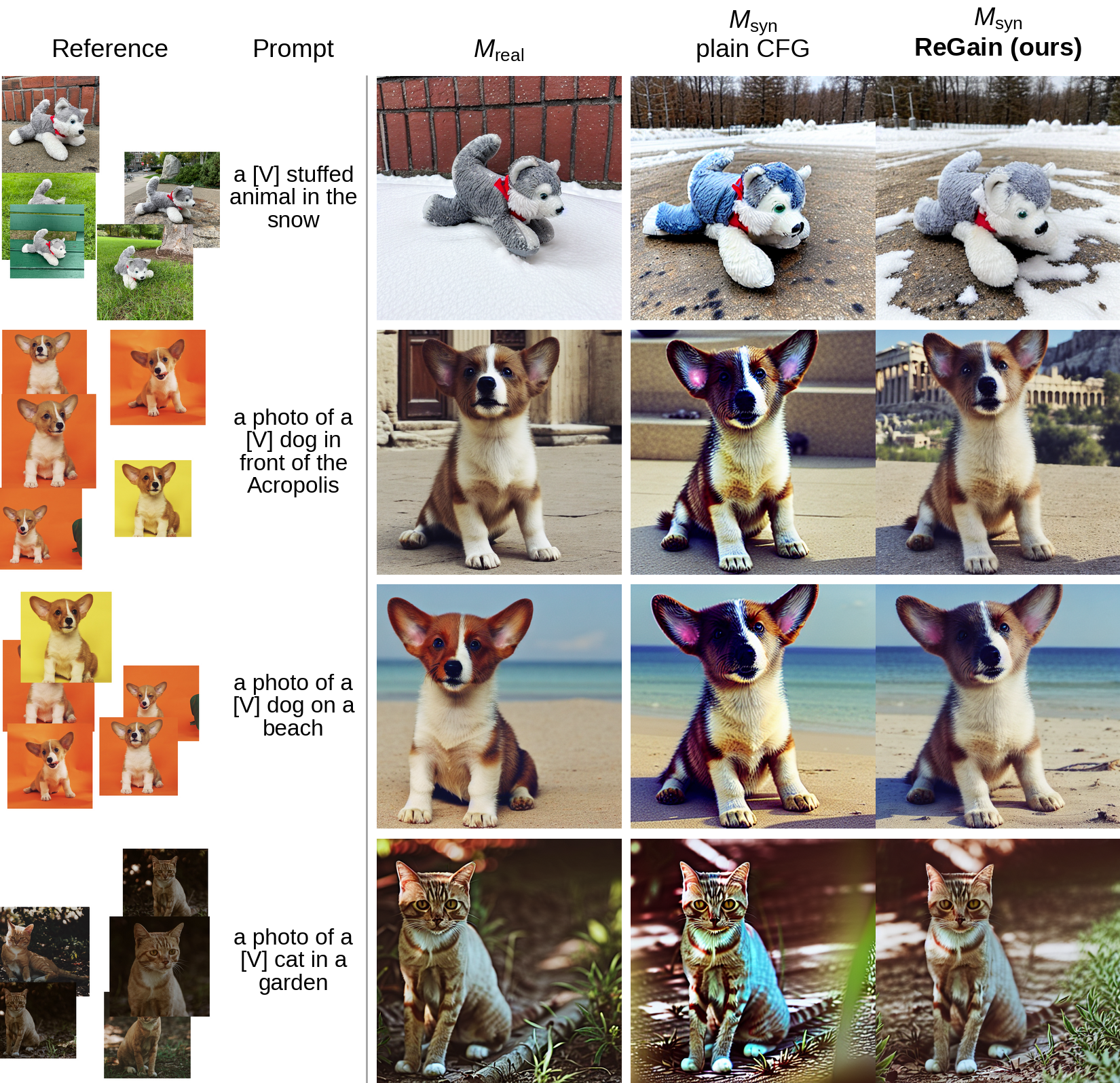}
\caption{\textbf{Additional subjects and prompts.} Four further subject and prompt
combinations, shown in the
same block at the same seed and guidance weight ($w{=}7.5$, $50$ steps): the
subject's reference photos, the prompt, then $\Mreal$, $\Msyn$ under plain CFG, and the
same $\Msyn$ under its estimated gain schedule (ReGain, ours).
}
\label{fig:app-subjects}
\end{figure}

\begin{figure}[p]
\centering
\includegraphics[width=0.7\linewidth]{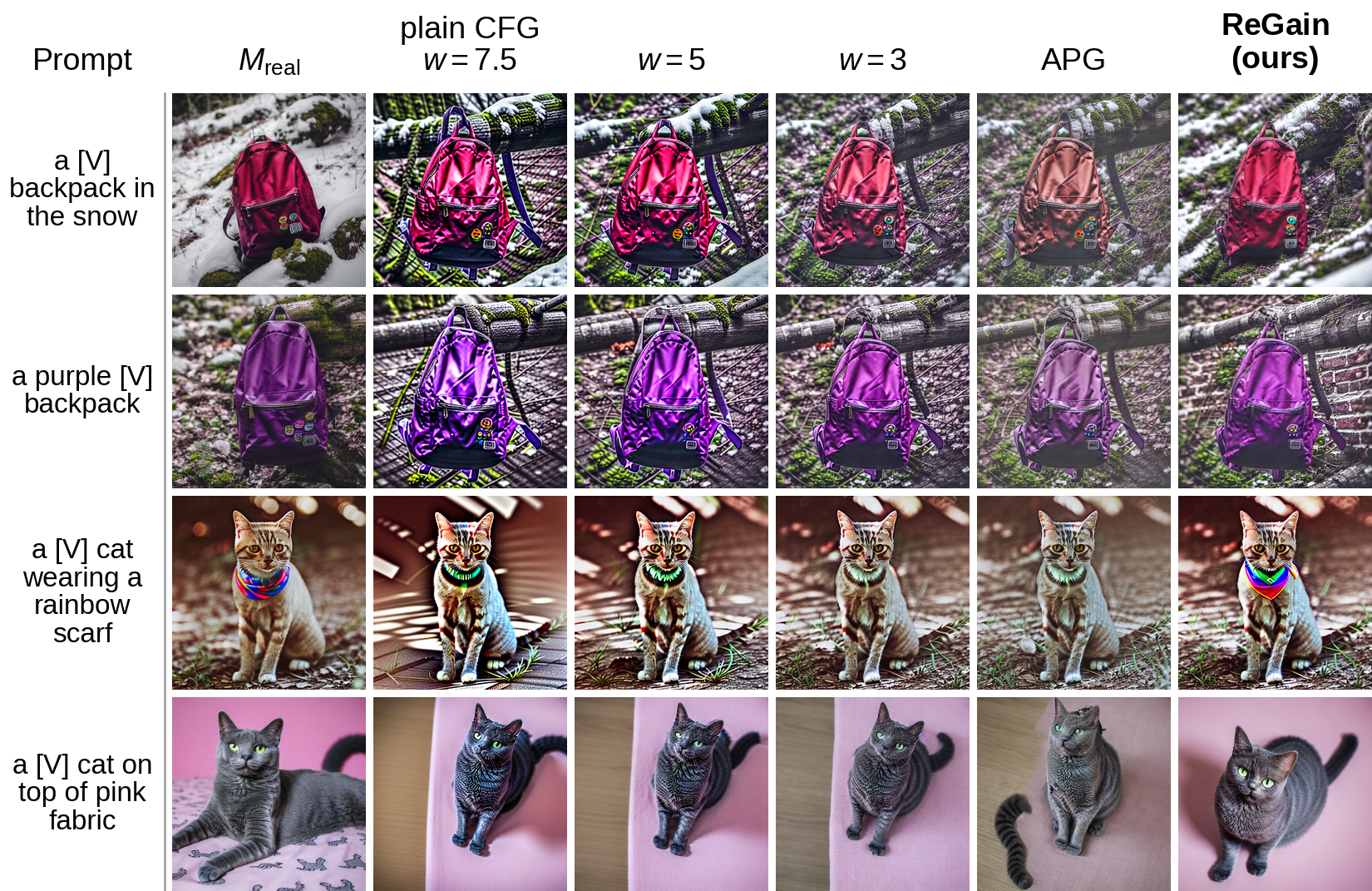}
\caption{\textbf{Subject-agnostic alternatives, further rows.} Four more
rows in the layout of Figure~\ref{fig:ablation-qual}, showing $\Mreal$, $\Msyn$ under
plain CFG at $w{=}7.5$, $5.0$ and $3.0$, under APG at $w{=}7.5$, and under
ReGain, at the same seed across columns.}
\label{fig:app-ablation}
\end{figure}

Figures~\ref{fig:app-lora-qual} and~\ref{fig:app-sdxl-qual} repeat the
comparison of Figure~\ref{fig:results-single} in the DreamBooth-LoRA setting on
Stable Diffusion v1.5 and SDXL, Figure~\ref{fig:app-subjects} adds further
subjects and prompts, and Figure~\ref{fig:app-ablation} adds further rows of
Figure~\ref{fig:ablation-qual}.

\end{document}